\documentclass{ppai}

 \usepackage{ppai}

\usepackage[utf8]{inputenc} % allow utf-8 input
\usepackage[T1]{fontenc}    % use 8-bit T1 fonts
\usepackage{hyperref}       % hyperlinks
\usepackage{url}            % simple URL typesetting
\usepackage{booktabs}       % professional-quality tables
\usepackage{amsfonts}       % blackboard math symbols
\usepackage{nicefrac}       % compact symbols for 1/2, etc.
\usepackage{microtype}      % microtypography
\usepackage{lipsum}
\usepackage{graphicx}
\graphicspath{ {./images/} }

\usepackage{xcolor}
\usepackage{pifont}
\usepackage{enumitem}
\usepackage{listings}
\usepackage{algorithm}
\usepackage{algpseudocode}
\usepackage{wrapfig}
\usepackage{subfig}
\usepackage{tcolorbox}

\usepackage{multirow}
\usepackage{amsmath}
\usepackage{amssymb}
\usepackage{mathtools}
\usepackage{amsthm}
\usepackage{dsfont}
\usepackage{array}
\usepackage{colortbl}  % For Table background color highlight

\usepackage[toc,page,header]{appendix}
\usepackage{minitoc}

\renewcommand{\partname}{}

\title{Entropy-Guided Attention for Private LLMs}

\author{
 Nandan Kumar Jha \\
  New York University \\
  \texttt{nj2049@nyu.edu} \\
   \And
 Brandon Reagen \\
   New York University \\
  \texttt{bjr5@nyu.edu} \\
}

\begin{document}

\doparttoc % Tell to minitoc to generate a toc for the parts
\faketableofcontents % Run a fake tableofcontents command for the partocs

%\part{} % Start the document part
%\parttoc % Use this only when you want to have ToC for main paper, too

\maketitle

\begin{abstract}
The pervasiveness of proprietary language models has raised critical privacy concerns, necessitating advancements in private inference (PI), where computations are performed directly on encrypted data without revealing users' sensitive information. While PI offers a promising solution, its practical deployment is hindered by substantial communication and latency overheads, primarily stemming from nonlinear operations. To address this, we introduce an information-theoretic framework to characterize the role of nonlinearities in decoder-only language models, laying a principled foundation for optimizing transformer-architectures tailored to the demands of PI.

By leveraging Shannon's entropy as a quantitative measure, we uncover the previously unexplored dual significance of nonlinearities: beyond ensuring training stability, they are crucial for maintaining attention head diversity. Specifically, we find that their removal triggers two critical failure modes: {\em entropy collapse} in deeper layers that destabilizes training, and {\em entropic overload} in earlier layers that leads to under-utilization of Multi-Head Attention's (MHA) representational capacity. 

We propose an entropy-guided attention mechanism paired with a novel entropy regularization technique to mitigate entropic overload. Additionally, we explore PI-friendly alternatives to layer normalization for preventing entropy collapse and stabilizing the training of LLMs with reduced-nonlinearities. Our study bridges the gap between information theory and architectural design, establishing entropy dynamics as a principled guide for developing efficient PI architectures. The code and implementation are available at \href{https://github.com/Nandan91/entropy-guided-attention-llm}{entropy-guided-llm}.

\end{abstract}

\section{Introduction}

The widespread deployment of proprietary large language models (LLMs) has raised critical privacy concerns for users' sensitive information \cite{staab2024beyond,mireshghallah2024can,priyanshu2023chatbots,ChatGptWiredArticle}. Private Inference (PI) offers a promising solution, enabling computations directly on encrypted data without exposing its contents. 

However, despite its potential, the practical deployment of PI systems remains a significant challenge due to substantial latency and communication overheads, particularly for transformer-based LLMs. Generating a single output token with a GPT-2 model (125M parameters) over 128 input tokens takes 8.2 minutes and requires 25.3 GB of communication (see Table \ref{tab:GPT2CLen128}). Scaling this to a context size of 512 results in 30.7 minutes and 145.2 GB of communication (Table \ref{tab:LanguiniGPT2}). 

These inefficiencies primarily arise from the computational overhead of nonlinear operations, which are critical for model stability and performance. Nonlinear computations in privacy-preserving settings require secure multi-party computation (MPC) protocols and cryptographic primitives such as secure comparisons, oblivious transfer, and polynomial evaluations (e.g., for GELU \cite{hendrycks2016gaussian}). These protocols involve multiple interaction rounds between users and service providers, significantly increasing communication and computational costs. 

For instance, a single GELU activation in a BERT-base model requires 3.9$\times 10^6$ point-wise operations, each involving multiple secure multiplications and communication rounds, typically adding 1-2 KB per operation \cite{lu2023bumblebee}. Recent work \cite{hou2023ciphergpt} has shown that nonlinear operations, primarily GELU and LayerNorm \cite{ba2016layer}, constitute the major bottleneck in PI, accounting for 49\% of latency and 59\% of communication costs. 

Designing LLMs with reduced-nonlinearities is a promising direction for efficient PI architectures. However, the fundamental role of nonlinearities in preserving transformer expressiveness and regulating internal information flow remains poorly understood. For instance,  Li et al. \cite{linonlinear} offer a theoretical analysis of attention and feed-forward network (FFN) nonlinearities in in-context learning tasks, which is limited to a simplified setting: a one-layer model with a single softmax-based self-attention head and a ReLU-based FFN.  While Cheng et al. \cite{cheng2023transformers} extend this investigation by analyzing a broader range of nonlinear architectures, they remain focused on specific in-context learning tasks. 

These findings, while valuable, do not {\em adequately} address the  comprehensive role of nonlinearities in maintaining model stability, and fostering attention head diversity in practical multi-layer LLMs or their implications for PI. Recent studies have shown an increasing focus on understanding the failure modes in transformer models, such as training instability \cite{wortsman2024smallscale,rybakov2024methods,zhai2023stabilizing} and rank collapse \cite{bao2024self,noci2022signal,dong2021attention}.  However, they predominantly focus on standard transformer architecture, leaving a critical question unaddressed: {\em How do the removal of non-linearities, impact training dynamics?}

To bridge this gap, we propose an information-theoretic framework to systematically analyze the role of nonlinearities in transformer-based models. Using Shannon's entropy as a quantitative lens, we uncover the {\bf dual significance} of nonlinearities: (1) they ensure training stability by preventing entropy collapse in deeper layers, and (2) they preserve the representational diversity of MHA by mitigating entropic overload in earlier layers,  fostering head-wise specialization.

{\bf Our Contributions:} Building on these insights, our work makes the following contributions: \vspace{-0.5em}
\begin{enumerate} [noitemsep,nolistsep,leftmargin=0.5cm] 
\item   {\em PI-friendly layer normalization alternatives:} To address training instability in LLM with reduced-nonlinearities, without relying on LayerNorm, we study the static normalization techniques such as weight and spectral normalization techniques \cite{salimans2016weight,miyato2018spectral}. These methods mitigate entropy collapse in deeper layers while avoiding the overheads associated with nonlinear operations in  LayerNorm.

\item {\em Entropy regularization techniques:} We introduce an entropy-guided attention mechanism and propose a novel entropy regularization technique to prevent entropic overload in LLMs with reduced-nonlinearities. Our approach incorporates two key innovations: (a) Headwise learnable thresholds to dynamically adjust regularization strength for each attention head, tailoring the process to the specific characteristic of individual heads; and (2) Tolerance margins to prevent over-regularization, preserving attention head diversity while preventing excessive penalization.

\item  {\em Practical design for PI:} We implement the entropy-guided framework and demonstrate their effectiveness across various context sizes (128, 256, 512) and model depths (12L and 18L) on a wide range of training tokens (1.2B to 4.8B) from the CodeParrot \cite{codeParrot} and Languini dataset \cite{stanic2023languini} on GPT-2 models.
\end{enumerate}

By analyzing entropy dynamics across layers, we provide a principled understanding of how architectural simplifications, such as removing nonlinearities, affect training stability and the representational diversity of attention heads in MHA. Our study establishes entropy dynamics as a foundational framework for optimizing privacy-preserving LLM architectures.

\section{Preliminaries}

{\bf Notations.}
We denote the number of layers as $L$, number of heads as $H$, model dimensionality as $d$, head dimension as $d_k$ (where \(d_k = \frac{d}{H}\)), and context length as $T$.  Table \ref{tab:ArchConfigGPT2} illustrates the abbreviations for architectural configurations with simplified nonlinearities in a transformer-based LLM.

{\bf An overview of transformer-based decoder-only architecture.}
A transformer-based LLM is constructed by sequentially stacking \(L\) transformer blocks, where each block is composed of two sub-blocks: an attention mechanism and a feed-forward network (FFN), both having their own residual connections and normalization layers, positioned in the Pre-LN order to improves training stability \cite{xiong2020layer}. Formally, transformer blocks take an input sequence \(\mathbf{X}_{\text{in}} \in \mathbb{R}^{T \times d}\), consisting of \(T\) tokens of dimension \(d\), and transform it into \(\mathbf{X}_{\text{out}}\) as follows:

\vspace{-1.5em}

\begin{equation} \label{eqn:ffn_mha}
\mathbf{X}_{\text{out}} = \hat{\mathbf{X}}_{\text{SA}} + \text{FFN}_{\text{GELU}}(\text{LayerNorm}_2(\hat{\mathbf{X}}_{\text{SA}})), \; \text{where} \; \hat{\mathbf{X}}_{\text{SA}} = \mathbf{X}_{\text{in}} + \text{MHA}(\text{LayerNorm}_1(\mathbf{X}_{\text{in}})).
\end{equation}

The Multi-Head Attention (MHA) sub-block enables input contextualization by sharing information between individual tokens. MHA employs the self-attention mechanism to compute the similarity score of each token with respect to all other tokens in the sequence, and transform the input sequence \(\mathbf{X}\) into  \(\mathbf{Attn}(\mathbf{X})\) as follows:

\vspace{-1.5em}

\begin{equation} \label{eqn:attn_softmax}
\mathbf{Attn}(\mathbf{X}) = \Big(\text{Softmax}\Big(\frac{1}{\sqrt{d_k}} (\mathbf{X} \mathbf{W}^Q) (\mathbf{X}{\mathbf{W}^K})^\top + \mathbf{M}\Big)\Big)\mathbf{X}\mathbf{W}^V.
\end{equation}

Here, each token generates query($Q$), key($K$), and value($V$) vectors through the linear transformations \(\mathbf{W}^Q, \mathbf{W}^K, \; \text{and} \; \mathbf{W}^V \in \mathbb{R}^{d \times d_h}\), respectively. Then, similarity scores are computed by taking the dot product of the $Q$ and $K$ vectors, scaled by the inverse square root of the $K$ dimension, and passed through a softmax function to obtain the attention weights. These weights are then used to compute a weighted sum of the $V$ vectors, producing the output for each token. For auto-regressive models (e.g., GPT), mask \(\mathbf{M} \in \mathbb{R}^{T\times T}\),  which has values in \(\{0, -\infty\}\) with \(\mathbf{M}_{i,j} = 0 \,\text{iff} \, {i \geq j}\), is deployed to prevent the tokens from obtaining information from future tokens.

The MHA sub-block employs a self-attention mechanism across all the heads, each with its own sets of $Q$, $K$, and $V$. This allows the attention heads to focus on different parts of the input sequence, capturing various aspects of the input data simultaneously. The outputs from all heads are concatenated and linearly transformed ($\mathbf{W}^O\in\mathbb{R}^{d\times d}$) to produce the final MHA output as follows: 

\vspace{-1.3em}

\begin{equation} \label{eqn:mha_concat}
\text{MHA}(\mathbf{X}) = \text{Concat}\big(\text{Attn}_1(\mathbf{X}), \; \text{Attn}_2(\mathbf{X}), \; \text{Attn}_3(\mathbf{X}), \dots, \text{Attn}_H(\mathbf{X})\big) \mathbf{W}^O.
\end{equation}

Following the MHA sub-block, the FFN sub-block transforms each token independently. The FFN sub-blocks have a single hidden layer whose dimension is a multiple of \(d\) (e.g., \(4d\) in GPT \cite{radford2019language}  models). The FFN sub-block first applies a linear transformation to the input \(\mathbf{X}\) using \(\mathbf{W}^{\text{ffn}}_{\text{in}} \in \mathbb{R}^{d \times 4d}\), followed by a non-linear transformation using an activation function such as GELU. This is then followed by another linear transformation using \(\mathbf{W}^{\text{ffn}}_{\text{out}} \in \mathbb{R}^{4d \times d}\), as follows:

\vspace{-1.5em}

\begin{equation} \label{eqn:ffn_gelu}
\text{FFN}(\mathbf{X}) = (\text{GELU}(\mathbf{X} \mathbf{W}^{\text{ffn}}_{\text{in}}))\mathbf{W}^{\text{ffn}}_{\text{out}}
\end{equation}

%The combination of MHA and FFN sub-blocks, along with residual connections and normalization layers, allows transformer models to  learn the contextual relationships between tokens effectively.

{\bf Threat model for private inference.} We consider the standard two-party (2PC) client-server setting used in PPML, which provides security against semi-honest (honest-but-curious) adversaries bounded by probabilistic polynomial time \cite{zhang2024secure,lu2023bumblebee,pang2023bolt,hou2023ciphergpt}.  Both parties follow protocol specifications but may attempt to gain additional information from their outputs about the other party's input. In this 2PC  setting, the server holds the propriety LLM (e.g., ChatGPT), and the client queries the model with a piece of text (prompt). The protocols ensure the server learns nothing about the client's input or query output, and the client learns nothing beyond the the server's model architecture.

\begin{figure} [htbp]
\centering
\includegraphics[width=1.0\textwidth]{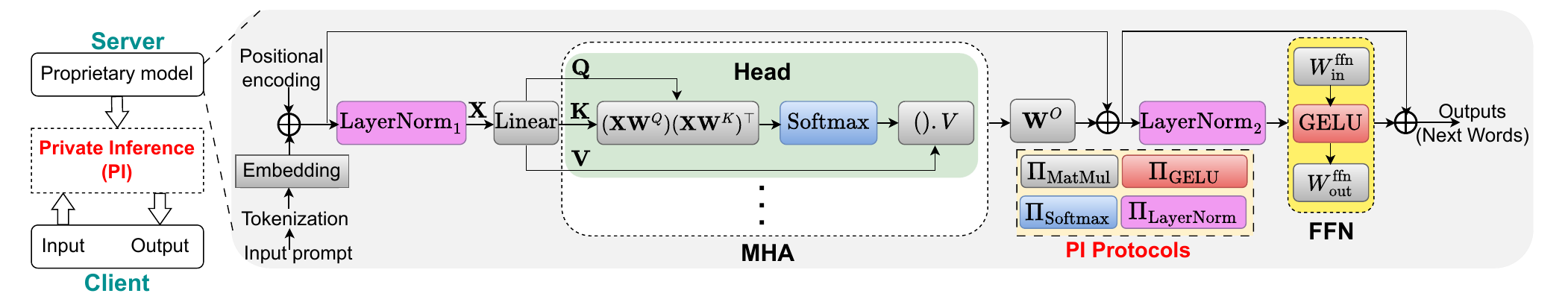} \vspace{-2.2em}
\caption{An illustration of threat model and cryptographic protocols used for LLM private inference.} 
\label{fig:GptBlockDiagram}
\end{figure}  

\vspace{-2em}

\begin{table}[htbp]
\caption{Architectural configurations of nonlinearities in LLMs, illustrating the combinations of Softmax (SM), LayerNorm (LN), GELU (G), and ReLU (R) functions (see Eq. \ref{eqn:ffn_mha}, \ref{eqn:attn_softmax}, \ref{eqn:mha_concat} and \ref{eqn:ffn_gelu}). } 
\label{tab:ArchConfigGPT2}
\centering 
\begin{tabular}{l|c} \toprule 
Abbreviation & Architectural configuration \\ \toprule 
\textcolor{blue}{SM} + \textcolor{violet}{LN} + \textcolor{red}{G} & $\mathbf{X}_{\text{out}} = \text{FFN}_{\text{\textcolor{red}{GELU}}}(\text{\textcolor{violet}{LayerNorm}}_{\textcolor{violet}{2}}(\text{MHA}(\text{Attn}_{\text{\textcolor{blue}{Softmax}}}(\text{\textcolor{violet}{LayerNorm}}_{\textcolor{violet}{1}}(\mathbf{X}_{\text{in}})))))$ \\
\textcolor{blue}{SM} + \textcolor{violet}{LN} + \textcolor{red}{R} & $\mathbf{X}_{\text{out}} = \text{FFN}_{\text{\textcolor{red}{ReLU}}}(\text{\textcolor{violet}{LayerNorm}}_{\textcolor{violet}{2}}(\text{MHA}(\text{Attn}_{\text{\textcolor{blue}{Softmax}}}(\text{\textcolor{violet}{LayerNorm}}_{\textcolor{violet}{1}}(\mathbf{X}_{\text{in}})))))$ \\ 
\textcolor{blue}{SM} + \textcolor{violet}{LN} & $\mathbf{X}_{\text{out}} = \text{FFN}_{\text{Identity}}(\text{\textcolor{violet}{LayerNorm}}_{\textcolor{violet}{2}}(\text{MHA}(\text{Attn}_{\text{\textcolor{blue}{Softmax}}}(\text{\textcolor{violet}{LayerNorm}}_{\textcolor{violet}{1}}(\mathbf{X}_{\text{in}})))))$ \\ 
\textcolor{blue}{SM} + \textcolor{red}{G} & $\mathbf{X}_{\text{out}} = \text{FFN}_{\text{\textcolor{red}{GELU}}}(\text{MHA}(\text{Attn}_{\text{\textcolor{blue}{Softmax}}}(\mathbf{X}_{\text{in}})))$ \\ 
\textcolor{blue}{SM} + \textcolor{red}{R} & $\mathbf{X}_{\text{out}} = \text{FFN}_{\text{\textcolor{red}{ReLU}}}(\text{MHA}(\text{Attn}_{\text{\textcolor{blue}{Softmax}}}(\mathbf{X}_{\text{in}})))$ \\ 
\textcolor{blue}{SM} & $\mathbf{X}_{\text{out}} = \text{FFN}_{\text{Identity}}(\text{MHA}(\text{Attn}_{\text{\textcolor{blue}{Softmax}}}(\mathbf{X}_{\text{in}})))$ \\ \bottomrule
\end{tabular} 
\end{table}

%{\bf Protocols used for linear layers:}

%\cite{zeng2023copriv,zeng2022mpcvit,kundu2023learning,cho2022selective,ghodsi2021circa,jha2021deepreduce}.
%\cite{zhang2024secure,lu2023bumblebee,zimerman2023converting,pang2023bolt,gupta2023sigma,hou2023ciphergpt}

%{\bf Protocols used for nonlinear layers:}

%{\bf FFN configurations: } When we eliminate the GELU activation layer from the FFN, we refer to it as {\em SlimGELU FFN}. Further, {\em FusedLayer FFN} refers to the FFN after consolidating the dual linear layers into a single layer post-GELU removal. Lastly, {\em IdentityLink FFN} characterizes a scenario where the FFN is simplified to an identity connection, essentially bypassing the network's transformation through FFN.

%where \(\mathbf{W}^Q, \mathbf{W}^K \in \mathbb{R}^{d \times d_k}\) and \(\mathbf{W}^V \in \mathbb{R}^{d \times d_v}\) are trainable query, key, and value parameters, respectively. Here, the attention matrix \(\mathbf{A}(\mathbf{X}) \in \mathbb{R}^{T \times T}\) can be thought of as allowing different tokens to "mix" with each other. \(\mathbf{M} \in \mathbb{R}^{T \times T}\) is a mask taking values in \(\{0, -\infty\}\) that depend on the modeling task. For causal auto-regressive transformers like GPT, \(\mathbf{M}_{i,j} = 0 \,\text{iff} \, {i \geq j}\), which prevents a token from obtaining information from future tokens. In bidirectional models like BERT, masking is typically applied at the token level and not in the attention mechanism (i.e., \(\mathbf{M}\) is the zero matrix).

\section{Information-Theoretic Analysis of Nonlinearity in LLMs}
In this section, we systematically decouple nonlinearities from transformer-based decoder-only LLMs, investigating their impact on training dynamics and expressiveness of attention mechanism, through the lens of Shannon's entropy.

{\bf Shannon's entropy for quantifying attention score distribution } 
Shannon's entropy quantifies the uncertainty in a probability distribution, measuring the amount of information needed to describe the state of a stochastic system \cite{shannon1948mathematical,jaynes1957information}. For a probability distribution \( P(x) \), the entropy is defined as \( \mathbf{E}(P) = -\sum_{i} P(x_i) \log P(x_i) \). 

In a softmax-based attention mechanism, each softmax operation yields an entropy value representing the sharpness or spread of the attention scores for each query position \cite{ghader2017does,vig2019analyzing}. Higher entropy indicates a more uniform distribution of softmax scores, while lower entropy signifies a more focused distribution on certain features \cite{nahshan2024linear}.

Let \(\mathbf{A}^{(h,l)} \in \mathbb{R}^{T\times T} \) be the attention matrix of $h$-th head in $l$-th layer, and each element in the attention matrix, \( a_{ij}^{(l, h)} \), are attention weights for the \(i\)-th query and \(j\)-th key, which are non-negative and sum to one for a query:

\vspace{-1.5em}

\begin{equation}
\mathbf{A}^{(l, h)} = \left[ a_{ij}^{(l, h)} \right]_{T \times T}, \quad \text{where} \quad a_{ij}^{(l, h)} \geq 0 \quad \text{and} \quad \sum_{j=1}^{T} a_{ij}^{(l, h)} = 1
\end{equation}

%\vspace{-1.5em}

This square matrix is generated by applying the softmax operation over the key length for each query position as follows 

\vspace{-1.5em}

\begin{equation} 
\mathbf{A}^{(h,l)}(\mathbf{X}) = \text{Softmax}\Big(\frac{1}{\sqrt{d_k}} (\mathbf{X} \mathbf{W}^Q) (\mathbf{X}{\mathbf{W}^K})^\top \Big), \; \text{where} \quad \text{Softmax}(\mathbf{X}_i) = \frac{\exp \left(x_{i} \right)}{\sum_{j=1}^{T} \exp \left( x_{j} \right)}
\end{equation}

\vspace{-1em}

Following \cite{zhai2023stabilizing}, we compute the mean of entropy values across all query positions to obtain a single entropy value for each head. The entropy \( \mathbf{E}^{(l, h)} \) for the \( h \)-th head in the \( l \)-th layer of an attention matrix is given by:

\vspace{-1.5em}

\begin{equation}
\mathbf{E}^{(l, h)} = -\frac{1}{T} \sum_{i=1}^{T} \sum_{j=1}^{T} a_{ij}^{(l, h)} \log \left( a_{ij}^{(l, h)} + \epsilon \right), \quad \text{where} \quad a_{ij}^{(l, h)} = \frac{\exp\left(\frac{1}{\sqrt{d_k}} (\mathbf{X}_i \mathbf{W}^Q) (\mathbf{X}_j \mathbf{W}^K)^\top \right)}{\sum_{k=1}^{T} \exp\left(\frac{1}{\sqrt{d_k}} (\mathbf{X}_i \mathbf{W}^Q) (\mathbf{X}_k \mathbf{W}^K)^\top \right)}
\end{equation}

\vspace{-1em}

where \( \epsilon \) is a small constant added for numerical stability to prevent taking the log of zero.

{\bf Well-behaved entropy distribution for LLMs} 
We begin by analyzing the headwise entropy distribution of baseline architecture with GELU (${\tt SM + LN + G}$) and ReLU (${\tt SM + LN + R}$) in their FFN. We find that the majority of heads ($\approx$90\%) possess entropy values between $\frac{\text{max}}{4}$ and $\frac{\text{3max}}{4}$, where ${\tt max}$ is maximum observed entropy value among all heads (see Figure \ref{fig:LossCurveNonlinConfig}a).  
This concentration in the mid-entropy range, while avoiding extremes, demonstrates a well-behaved distribution, providing a benchmark for assessing the impact of nonlinearities on model behavior.

{\bf Entropic overload in nonlinearity-reduced LLMs}
We observed that in certain nonlinearity configurations, a disproportionately large fraction of the attention heads exhibit higher entropy values (between $\frac{3\text{max}}{4}$ and ${\tt max}$), and we term this phenomenon as entropic overload. We hypothesize that this deviation form well-behaved entropy distribution results in {\em under-utilization} of the network's representational capacity, as too many heads engaged in exploration, hindering the model from effectively leveraging the diversity of attention heads. 

\vspace{-1.5em}

\begin{table}[htbp]
\begin{minipage}[b]{0.71\linewidth}
\centering
\subfloat[Headwise entropy distribution]{\includegraphics[width=.5\textwidth]{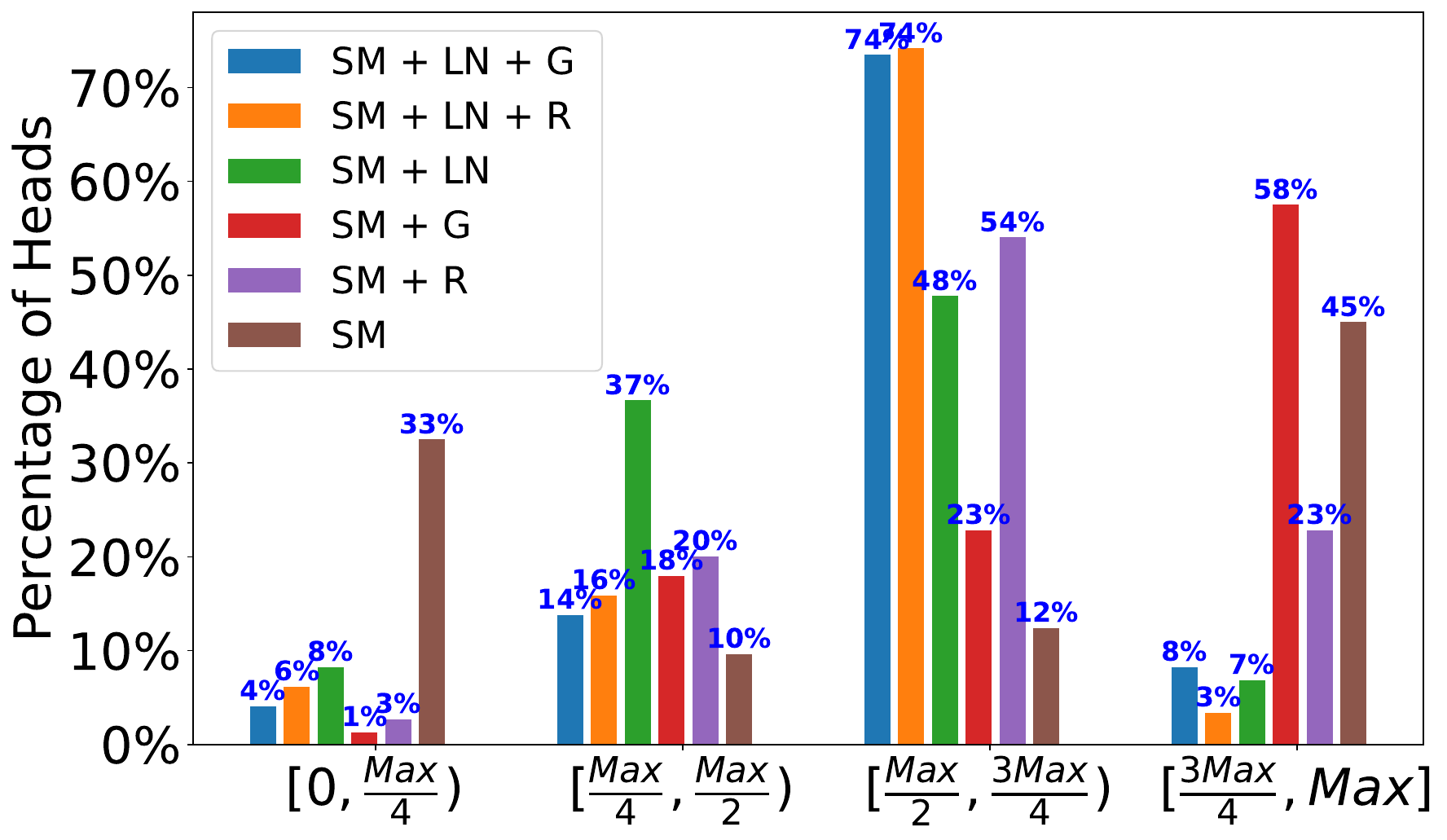}}
\subfloat[Loss curve]{\includegraphics[width=.5\textwidth]{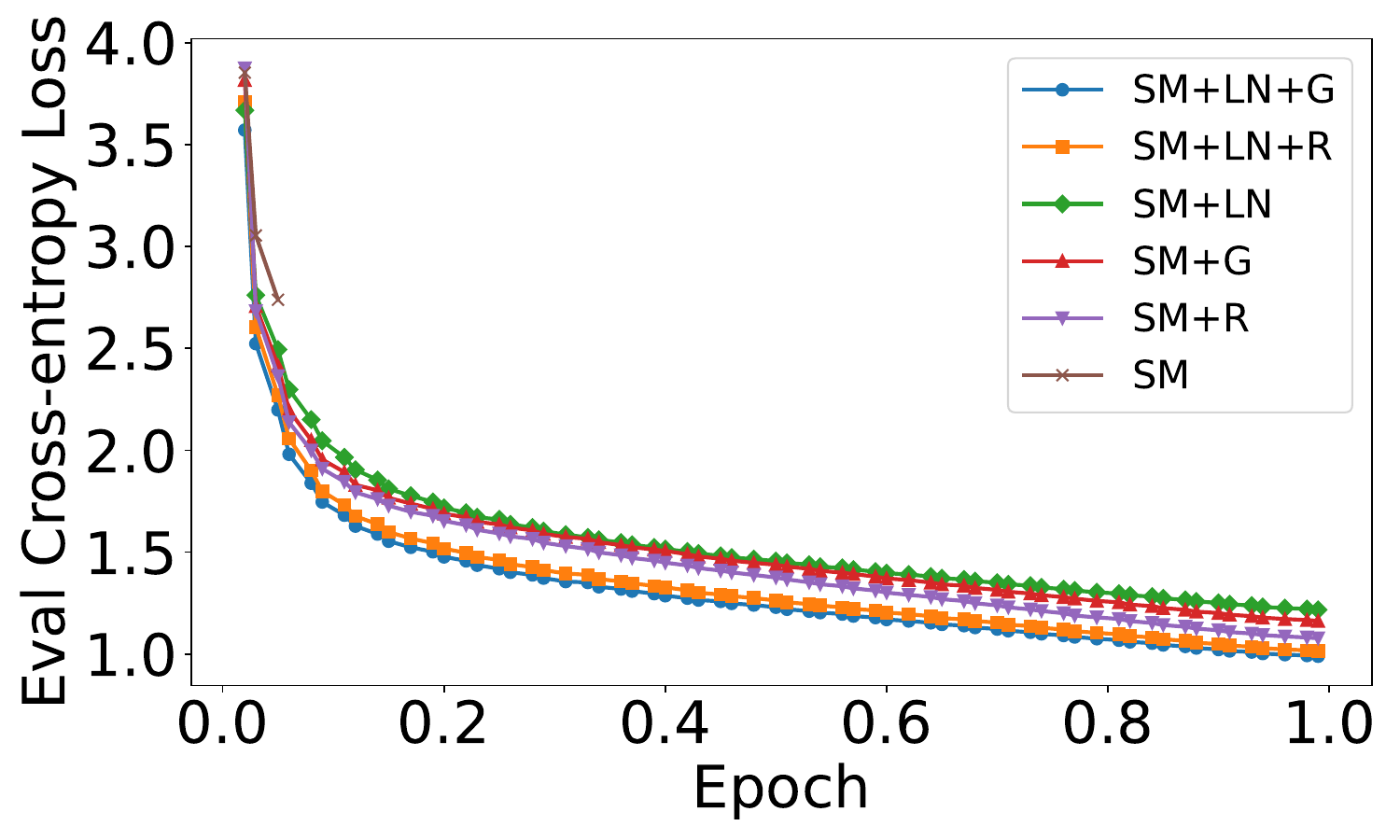}} 
\captionof{figure}{ (a) The  fraction of attention heads distributed across different entropy ranges, and (b) evaluation loss for GPT-2 (small) models with reduced-nonlinearities, when trained from scratch on CodeParrot dataset.}
\label{fig:LossCurveNonlinConfig}
\end{minipage} \hfill
\begin{minipage}[b]{0.27\linewidth}
\centering 
\resizebox{0.99\textwidth}{!}{
\begin{tabular}{lcl} \toprule
Configurations & PPL & +$\Delta$(\%)\\ \toprule
SM + LN + G & 2.69 & 0.00  \\
SM + LN + R & 2.76 & 2.53  \\
SM + LN & 3.38 &     25.58     \\
SM + G & 3.20 &      18.92     \\
SM + R & 2.94 &      9.20      \\
SM & NaNs &       -          \\  \bottomrule 
\end{tabular}}   
\captionof{table}{Evaluation perplexity for GPT-2 (small) models with reduced-nonlinearities, corresponding to Figure \ref{fig:LossCurveNonlinConfig}b. $\Delta$ is increase in eval PPL over baseline network.}
\label{tab:NonlinConfigEvalPPL}
\end{minipage}
\end{table}
%\vspace{-2em}

\vspace{-1.5em}

\begin{figure} [t]
\centering
%\subfloat[Layerwise entropy values]{\includegraphics[width=1.0\textwidth]{plots/layerwise_mean_ent_no_ln}} \\ \vspace{-0.85em}
%\subfloat[Layerwise entropy variance]{\includegraphics[width=.99\textwidth]{plots/layerwise_variance_entropy_no_ln}}   \\ \vspace{-0.85em}
\subfloat[SM + LN + G \label{subfig:sm_ln_g_hmap}]{\includegraphics[width=.3\textwidth]{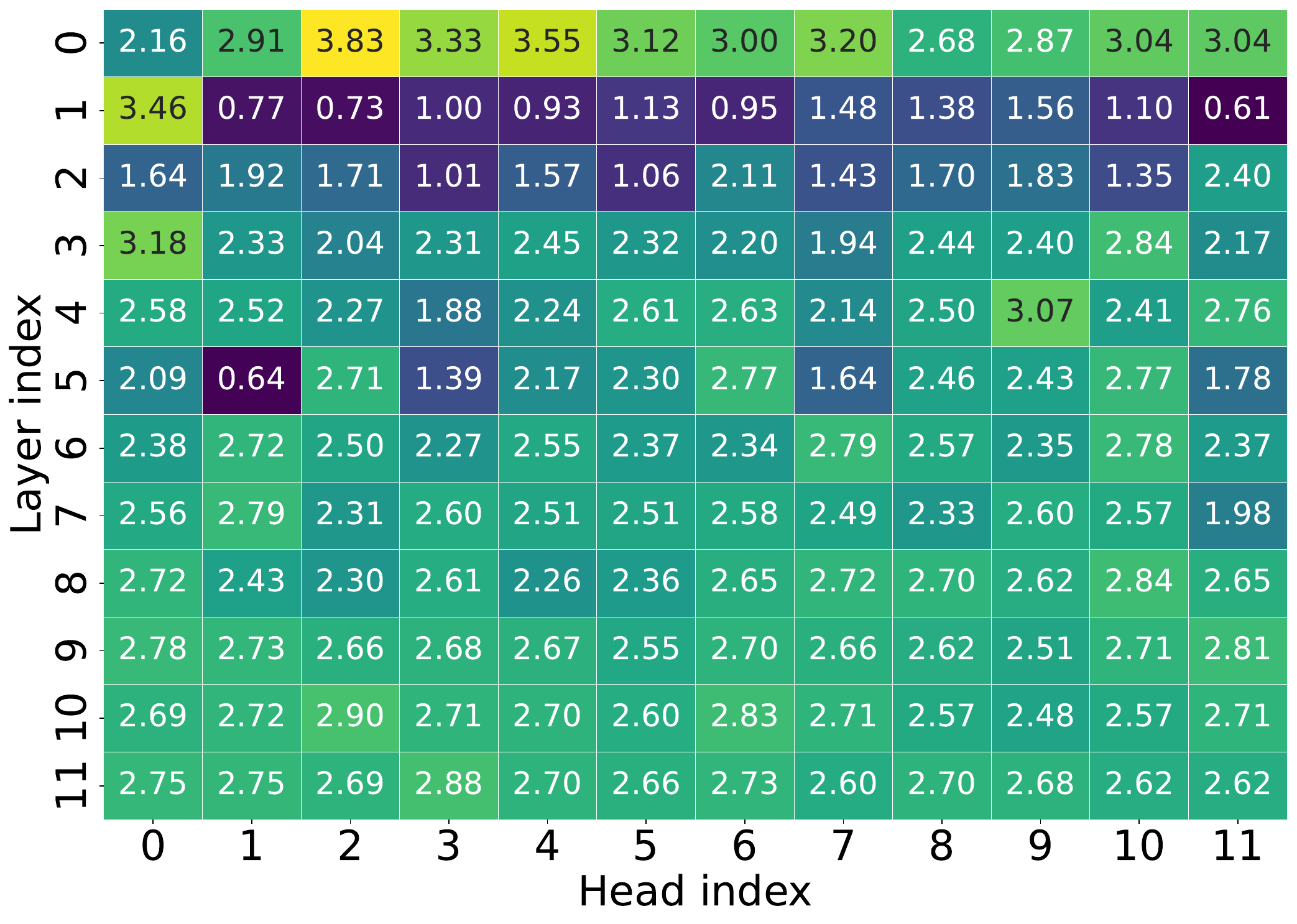}}
\subfloat[SM + LN + R \label{subfig:sm_ln_r_hmap}]{\includegraphics[width=.3\textwidth]{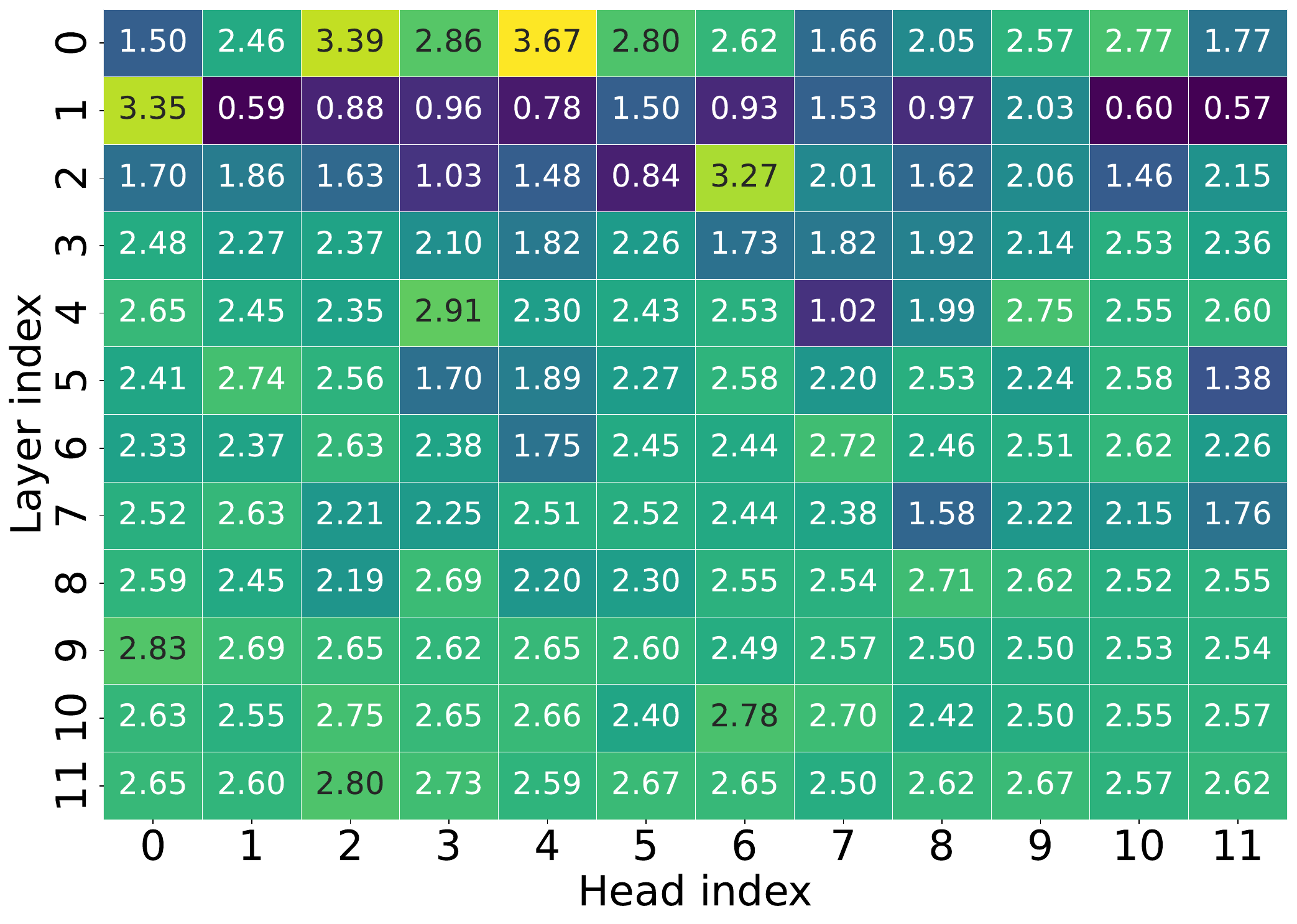}} 
\subfloat[SM + LN \label{subfig:sm_ln_hmap}]{\includegraphics[width=.3\textwidth]{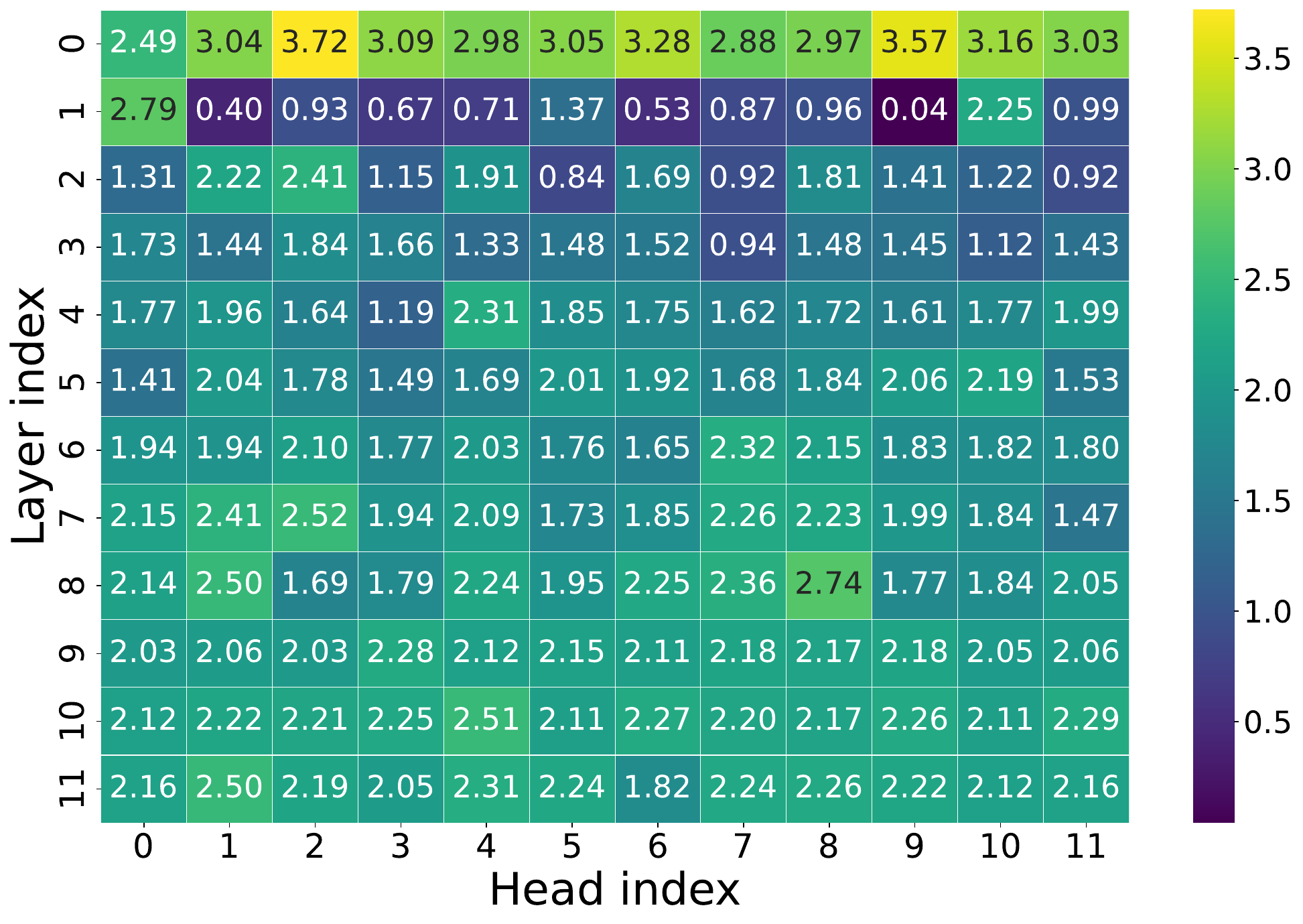}} \\ \vspace{-1em}
\subfloat[SM + G \label{subfig:sm_g_hmap}]{\includegraphics[width=.3\textwidth]{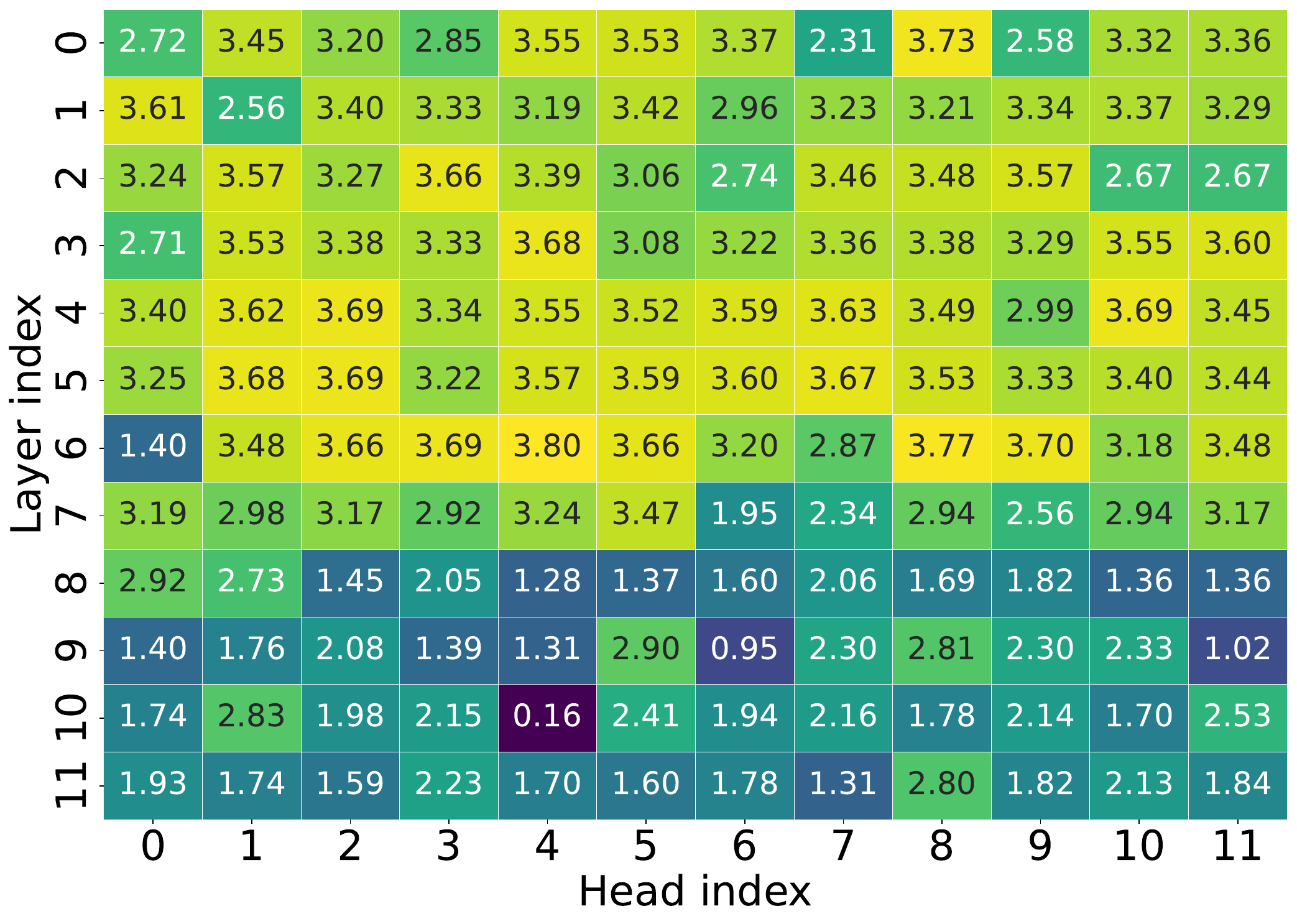}} 
\subfloat[SM + R \label{subfig:sm_r_hmap}]{\includegraphics[width=.3\textwidth]{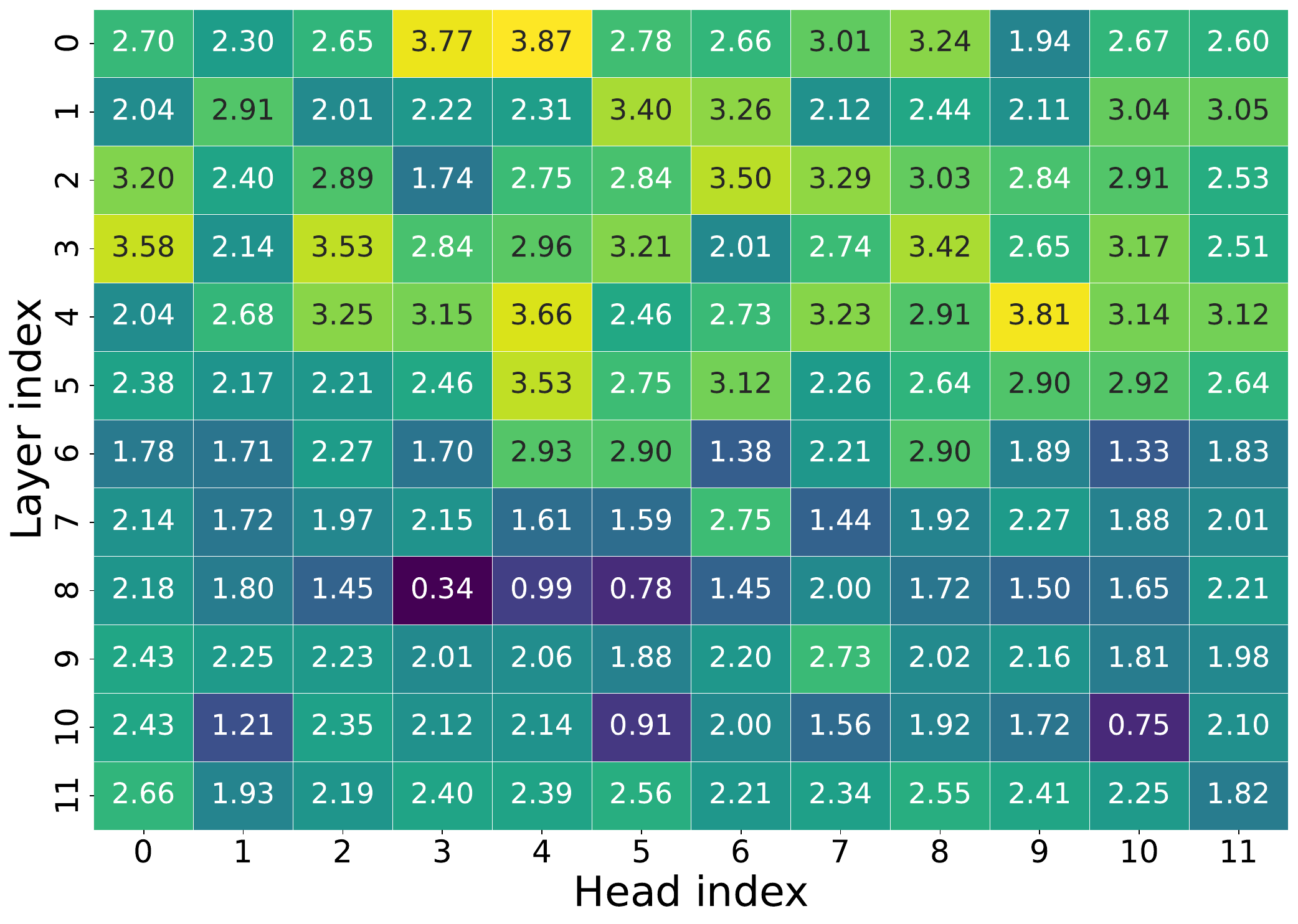}} 
\subfloat[ SM \label{subfig:sm_only_hmap}]{\includegraphics[width=.3\textwidth]{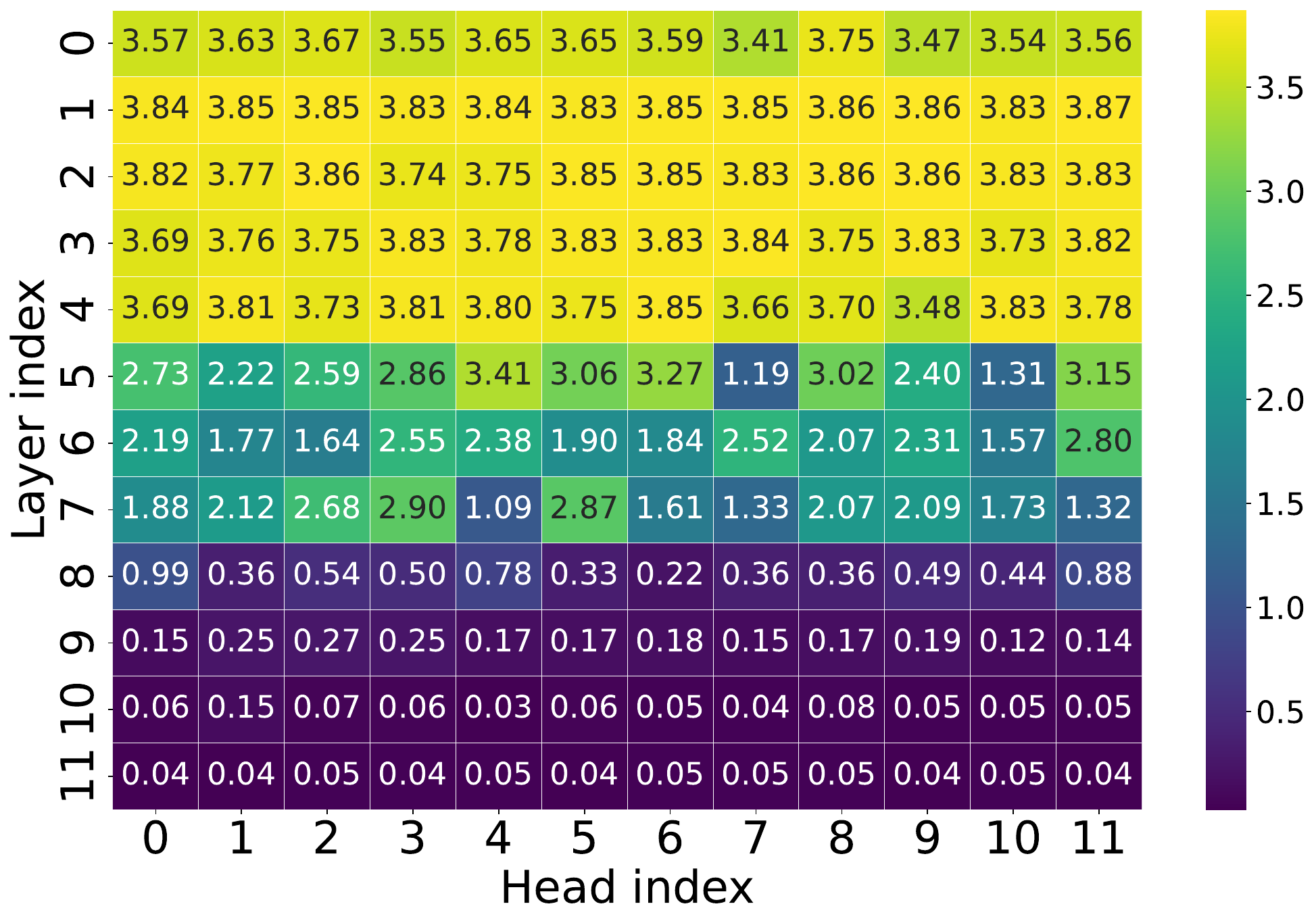}} 
\vspace{-0.8em}
\caption{Headwise entropy distribution in LLM architectures with reduced nonlinearities compared to baseline models. Yellow regions indicate high-entropy concentrations, revealing severe entropic overload predominantly in early layers. }
\label{fig:AttnEntHeatMaps}
\end{figure}

We visualize the entropy heatmaps for LLM architectures with reduced nonlinearity, trained from scratch (Figure \ref{fig:AttnEntHeatMaps}). Our analysis reveals severe entropic overload in the early layers of two specific architectures: the LayerNorm-free model with GELU (Figure \ref{subfig:sm_g_hmap}) and the Softmax-only model without LayerNorm and FFN activations (Figure \ref{subfig:sm_only_hmap}). 

Specifically, 58\% of heads in the LayerNorm-free GELU model have entropy values in the range $\frac{\text{3max}}{4}$ and ${\tt max}$, compared to only 23\% in the LayerNorm-free ReLU model (Figure \ref{fig:LossCurveNonlinConfig}a). Additionally, very few heads in the latter approach maximum entropy, unlike their GELU counterpart (see yellow regions in Figure \ref{subfig:sm_r_hmap} and Figure \ref{subfig:sm_g_hmap}), which results in 8.2\% improvement in perplexity (see Table \ref{tab:NonlinConfigEvalPPL}). On the other hand, the 45\% of heads in Softmax-only model have entropy values in the $\frac{\text{3max}}{4}$ to ${\tt max}$ range, with many approaching the maximum (Figure \ref{subfig:sm_only_hmap}).

{\bf Entropy collapse in nonlinearity-reduced LLMs}
The absence of LayerNorm and FFN nonlinearity in Softmax-only model leads to entropy collapse in the deeper layers---a phenomenon characterized by near-zero entropy values and recognized as a key indicator of training instability in transformer architectures \cite{zhai2023stabilizing,he2024understanding}. Quantitatively, 33\% of attention heads demonstrate entropy values within the range of 0 to $\frac{\text{max}}{4}$ (Figure \ref{fig:LossCurveNonlinConfig}a), with a significant concentration approaching zero (see Figure \ref{subfig:sm_only_hmap}). This systematic entropy collapse directly contributes to training instability, highlighting the critical role of nonlinear components in maintaining stable training dynamics.

\section{Entropy-Guided LLM Architecture for Efficient Private Inference}
We begin by exploring PI-friendly techniques to prevent entropy collapse in the absence of LayerNorm and FFN activations. Subsequently, we introduce an entropy-guided attention mechanism paired with an entropy regularization technique to mitigate entropic overload in nonlinearity-reduced LLMs.

{\bf PI-friendly layer normalization alternatives}
To address training instability, prior work has predominantly relied on LayerNorm applied to various parts of the network, such as QK-LayerNorm \cite{dehghani2023scaling,wortsman2024smallscale,muennighoff2024olmoe} and FFN-LayerNorm \cite{rybakov2024methods}. Since LayerNorm requires expensive inverse-square-root operations during inference \cite{hou2023ciphergpt},  we shift our focus from activation normalization to weight normalization techniques that avoid nonlinear computations at inference. 

We discover that the weight normalization \cite{salimans2016weight} and spectral normalization \cite{miyato2018spectral} serves as static alternatives to LayerNorm by normalizing weights instead of activations. These methods incur no additional cost at inference, and effectively prevent entropy collapse in the deeper layers of LLMs, in the absence of LayerNorm and FFN activations (see Figure \ref{fig:LayerwiseEntropyWSNorm}). Notably, the effectiveness of weight and spectral normalization  depends on targeting the appropriate linear layers, as applying them in attention sub-block {\em diminishes} overall performance compared to when applied in FFN (see Table \ref{tab:SNormVsWNorm}).

Furthermore, we employ a simpler technique to scale the outputs of the FFN sub-block by having learnable scaling factors for the FFN output and their residual output as follows (see Eq. \ref{eqn:ffn_mha}): 

\vspace{-1.5em}
\begin{equation} \label{eqn:ScaledFFN}
\mathbf{X}_{\text{out}} = \beta\hat{\mathbf{X}}_{\text{SA}} +\frac{1}{\alpha} (\text{FFN}^{\text{SM}}(\mathbf{X}_{\text{SA}})) \quad \text{where} \quad \alpha, \beta \in \mathbb{R}^{L}
\end{equation}

\vspace{-0.8em}

%Finally, we incorporate learnable FFN scaling into LLM architectures without LayerNorm and GELU activations, achieving lower perplexity and outperforming weight and spectral normalization (see Table \ref{tab:SNormVsWNormVsMlpGains}).

{\bf Architectural simplifications and entropy-guided attention} 
%\begin{figure} [t]
%\centering
%\includegraphics[width=.995\textwidth]{plots/GPT_Arch-AERO} \vspace{-2em}
%\caption{Overview of the proposed AERO method for reducing nonlinearities and FLOPs in transformer-based LLMs for efficient PI. The bottom of the figure shows the evolution of entropy in the attention mechanism and its distribution across attention heads.} 
%\label{fig:EndToEndAERO}
%\end{figure} 

\begin{wrapfigure}[17]{r}{0.48\textwidth - .95\columnsep}
\centering
\vspace{-1.9\intextsep}
\includegraphics[scale=0.47]{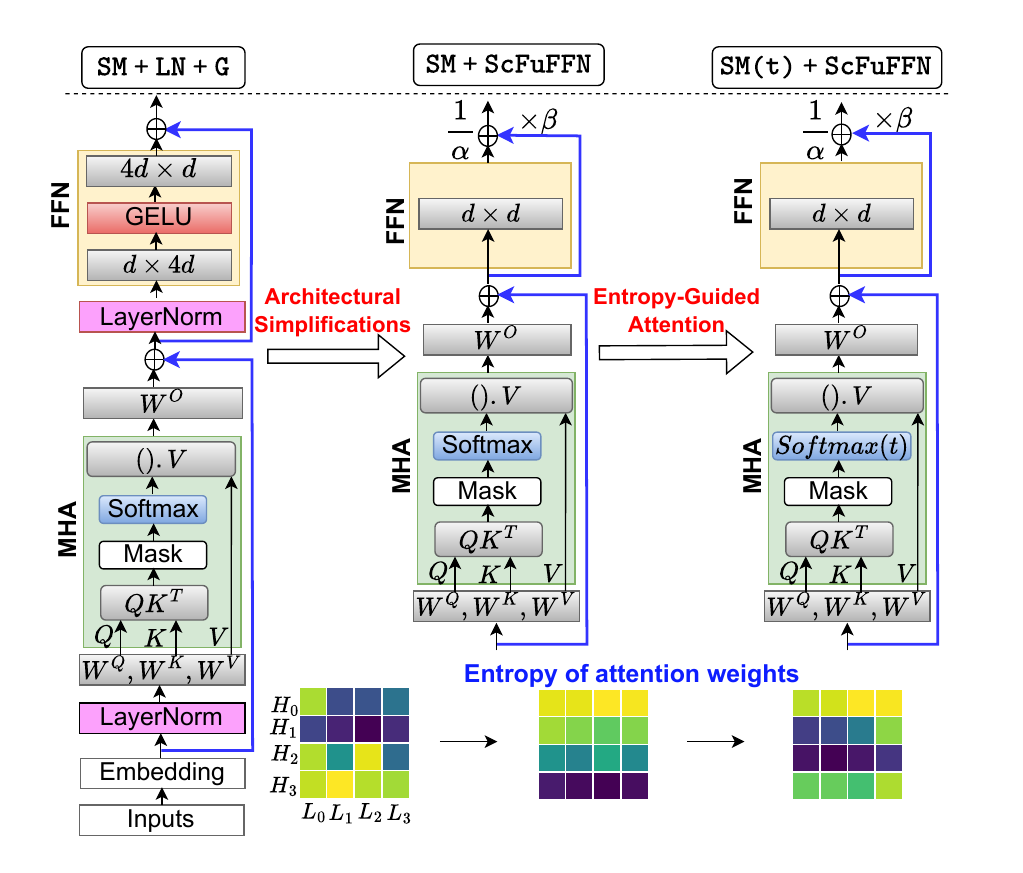} 
\vspace{-2.2em}
\caption{Nonlinearity-reduced simplified architecture  with entropy-guided attention mechanism. 
%The bottom of the figure shows the evolution of entropy in the attention mechanism.
} 
\label{fig:EndToEndAERO}
\end{wrapfigure}

We simplified the LLM architecture by designing a Softmax-only model that eliminates LayerNorm and FFN nonlinearity. Subsequently, we merge the two linear layers in the FFN---$\mathbf{W}^{\text{ffn}}_{\text{in}} \in \mathbb{R}^{d \times 4d}$ and $\mathbf{W}^{\text{ffn}}_{\text{out}} \in \mathbb{R}^{4d \times d}$---into a single linear layer $\mathbf{W}^{\text{ffn}} \in \mathbb{R}^{d \times d}$ (see Figure \ref{fig:EndToEndAERO}), as they perform equivalent linear transformations in the absence of intervening nonlinearity. However, training this simplified LLM presents challenges, particularly entropy collapse in deeper layers. To address this, we incorporate FFN scaling method that employ learnable scaling factors $\alpha$ and $\beta$ in the FFN sub-block. This approach stabilizes training more effectively than weight or spectral normalization, achieving lower perplexity (Table \ref{tab:SNormVsWNormVsMlpGains}). We denote this simplified model as $\texttt{SM+ScFuFFN}$ (Figure \ref{fig:EndToEndAERO}).

To preserve attention head diversity in our simplified architecture, we develop an entropy-guided attention mechanism. Inspired by \cite{miller1996global}, which employed temperature as a Lagrangian multiplier to control stochastic system entropy, we augment $\texttt{SM+ScFuFFN}$ with learnable temperatures for each softmax operation ($t \in \mathbb{R}^{H \times T}$). This allows the model to dynamically adjust entropy patterns during training by adjusting the temperature. Specifically, higher temperature values ($t > 1$) diffuse attention scores and increase entropy,  while lower values ($t < 1$) sharpen attention scores and reduce entropy (see Appendix~\ref{subsecAppendix:EntRegFullDerivation}). We refer this  simplified architecture with entropy-guided attention as $\texttt{SM(t)+ScFuFFN}$.

{\bf Design principles for entropy regularization schemes to prevent entropic overload.}  
Prior entropy regularization approaches have primarily aimed at penalizing low-entropy predictions \cite{setlur2022maximizing, pereyra2017regularizing}, based on the principle of maximum entropy \cite{jaynes1982rationale}. However, our goal is to regularize higher entropy values, which presents two-fold challenges: (1) Since each attention head captures different aspects of the input, the regularization strength needs to be adjusted for each head individually. (2) Some heads naturally exhibit higher entropy even in well-behaved entropy distributions, thus, penalizing all high-entropy values without distinction could be harmful, requiring a more flexible approach.

Followings are the key design principles for our entropy regularization scheme (see Algorithm \ref{Algo:EntRegLossComputation}):

\begin{itemize}[noitemsep,nolistsep,leftmargin=0.3cm] \vspace{-0.5em}
\item {\em Dynamic thresholds with head-specific adaptation: } To adapt the regularization strength based on the characteristics of each attention head \cite{voita2019analyzing},  we use headwise learnable threshold parameter $\mathtt{reg\_threshold\_weights} \in \mathbb{R}^H$. Consequently, the threshold for each head is computed as a learnable fraction of the maximum value of entropy ($\mathtt{reg\_threshold\_weights} \times \text{E}_{\text{max}}$), providing the fine-grained control (see Algorithm \ref{Algo:EntRegLossComputation}, line \#\ref{line:LearnableRegThreshold}).

\item {\em Tolerance margin to prevent over-regularization: }
To prevent over-regularization, we allow small deviations from the respective thresholds. Thus, a penalty is imposed only if the deviation from the threshold exceeds the tolerance margin, which is set as a fraction of \(\text{E}_{\text{max}}\) using the hyper-parameter \(\gamma\) (see Algorithm\ref{Algo:EntRegLossComputation}, line \#\ref{line:ToleranceMargin}). 

\vspace{-1em}

\begin{equation*} \label{eqn:EntThersDeviation}
\text{penalty}^{(l,h)} = 
\begin{cases} 
\Big( \text{deviation}^{(l,h)} \Big)^2 & \text{if } \big| \text{deviation}^{(l,h)} \big| > \gamma E_{\text{max}} \\ 
0 & \text{otherwise}
\end{cases} 
\end{equation*}

The deviation from threshold is computed as $\text{deviation}^{(l,h)} = \text{E}^{(l,h)}(t) -  \theta^{(l,h)} \text{E}_{\text{max}}$, where $\theta^{(l,h)}$ is $\mathtt{reg\_threshold\_weights}$. The hyper-parameter \(\gamma\) ensures that the model is not excessively penalized for minor deviations from the desired entropy threshold, which could impede its capacity to learn effectively. This careful calibration between stringent regularization and desired flexibility improves the model's robustness while maintaining its adaptability to various input distributions.

\item {\em Maximum entropy reference:} We set \(E_{\text{max}} = \log(T)\) as a reference point for computing thresholds and tolerance margins to ensure consistency across different layers and heads for regularization. Specifically, it provides a quantifiable reference for measuring deviations in entropy, making the regularization process more understandable.

%\item {\em Ease of integration:} Finally, integrating this entropy regularization with standard loss functions is straightforward ($\mathcal{L}_{\text{total}} = \mathcal{L}_{\text{CE}} + \lambda \mathcal{L}_{\text{entropy}}$), where \(\lambda\) controls the regularization strength. This hyper-parameter allows for flexible tuning based on specific needs without significant changes to existing training pipelines.

\end{itemize}

\begin{algorithm} 
\caption{Entropy Regularization Loss Computation} \label{Algo:EntRegLossComputation}
\textbf{Inputs:} $\text{attentions}$: List of attention matrices,  $\Theta(L,H)$= $\text{reg\_threshold\_weights}$, $T$: Sequence length, $\lambda$: Regularization loss weightage, $\gamma$: Hyper-parameter for  Tolerance margin\\
\textbf{Output:} $\mathcal{L}_{\text{total}}$: Total loss including entropy regularization
\begin{algorithmic}[1] 
\State $\mathcal{L}_{\text{entropy}} \leftarrow 0$
\State $\text{E}_{\text{max}} \leftarrow \log(T)$ \Comment{Theoretical maximum value of entropy}
\State \(\text{Tol}_{\text{margin}} \leftarrow \gamma \text{E}_{\text{max}} \) \Comment{Tolerance margin is set as a small fraction of $\text{E}_{\text{max}}$} \label{line:ToleranceMargin}
%\State $H \leftarrow \text{block.attn.num\_heads}$ 
%\State $L \leftarrow \text{len(attentions)}$ 
\For{each layer $l$ in layers}
    \State $\mathcal{L}_{\text{layer}} \leftarrow 0$
    \State $\text{A}(t) \leftarrow \text{attentions}[l]$  \Comment{Attention matrix with learnable temperature for each query position}
    %\State $\text{block} \leftarrow \text{blocks}[l]$
    \State $\text{E}(t) \leftarrow -\frac{1}{T} \sum_{i=1}^{T} \sum_{j=1}^{T} \text{A}_{ij}(t) \log(\text{A}_{ij}(t))$  \Comment{Compute entropy, averaged over query length}
    \For{each head $h$ in heads}
     \State $E^{(l,h)} \leftarrow \text{Slice}(\text{E}(t), h)$  \Comment{Entropy for head $h$}
            \State $\theta^{(l, h)} \leftarrow  \text{Slice}(\Theta{(L, H), h}) $  \Comment{Learnable threshold weight head $h$} 
     \State $\delta^{(l,h)} \leftarrow \text{E}^{(l,h)}(t) - \theta^{(l,h)} \text{E}_{\text{max}}$  \Comment{Deviation from head-specific threshold} \label{line:LearnableRegThreshold}
        \State $\text{penalty}^{(l,h)} \leftarrow (\delta^{(l,h)})^2 \mathds{1}(|\delta^{(l,h)}| > \text{Tol}_{\text{margin}})$  \Comment{Penalize iff deviation exceeds Tolerance}
        \State $\mathcal{L}_{\text{layer}} \leftarrow \mathcal{L}_{\text{layer}} + \text{penalty}^{(l,h)}$
    \EndFor
    \State $\mathcal{L}_{\text{layer}} \leftarrow \frac{\mathcal{L}_{\text{layer}}}{\text{num\_heads}}$  \Comment{Average over heads}
    \State $\mathcal{L}_{\text{entropy}} \leftarrow \mathcal{L}_{\text{entropy}} + \mathcal{L}_{\text{layer}}$
\EndFor
\State $\mathcal{L}_{\text{entropy}} \leftarrow \frac{\mathcal{L}_{\text{entropy}}}{\text{len(attentions)}}$  \Comment{Average over layers}
\State $\mathcal{L}_{\text{total}} \leftarrow \mathcal{L}_{\text{CE}} + \lambda \mathcal{L}_{\text{entropy}}$ \\
\Return $\mathcal{L}_{\text{total}}$
\end{algorithmic}
\end{algorithm}

\section{Experimental Results}

{\bf System setup}
We use a SecretFlow setup \cite{lu2023bumblebee} with the client and server simulated on two physically separate machines, each equipped with an AMD EPYC 7502 server with specifications of 2.5 GHz, 32 cores, and 256 GB RAM. We measure the {\em end-to-end} PI latency, including input embeddings and final output (vocabulary projection) layers, in WAN setting (bandwidth:100Mbps, latency:80ms),  simulated using Linux Traffic Control (tc) commands. The number of threads is set to 32. Following \cite{he2024simplifying,stanic2023languini,geiping2023cramming}, all the models are trained on a single RTX 3090 GPU.

{\bf Models and datasets} We train GPT-2 (12 and 18 layers) models on the CodeParrot \cite{codeParrot} and Languini book \cite{stanic2023languini} datasets, which are standard benchmarks for LLMs \cite{he2024simplifying,he2024understanding}. The CodeParrot dataset, sourced from 20 million Python files on GitHub, contains 8 GB of files with 16.7 million examples, each with 128 tokens, totaling 2.1 billion training tokens. We use a tokenizer with a vocabulary of 50K and train with context lengths of 128 and 256. The Languini book dataset includes 84.5 GB of text from 158,577 books, totaling 23.9 billion tokens with a WikiText-trained vocabulary of 16,384, and train with context length of 512. Each book averages 559 KB of text or about 150K tokens, with a median size of 476 KB or 128K tokens. 

{\bf Training Hyperparameters}
For pre-training on the CodeParrot dataset, we adopt the training settings from \cite{he2024simplifying}. Similarly, for training on the Languini dataset, we follow the settings from \cite{stanic2023languini}. These settings remain consistent across all architectural variations to accurately reflect the impact of the architectural changes. When applying entropy regularization on the CodeParrot dataset, we initialize the learnable temperature to 1e-2 and set $\lambda$ to 1e-5. For the Languini dataset, the temperature is initialized to 1e-1, and $\lambda$ is set to 5e-5.

%\vspace{-0.5em} \input{figures/6_fig_entropic_overlaod_solved}

\begin{figure} [htbp]
\centering
\subfloat[SM + LN + G \label{subfig:sm_ln_g}]{\includegraphics[width=.33\textwidth]{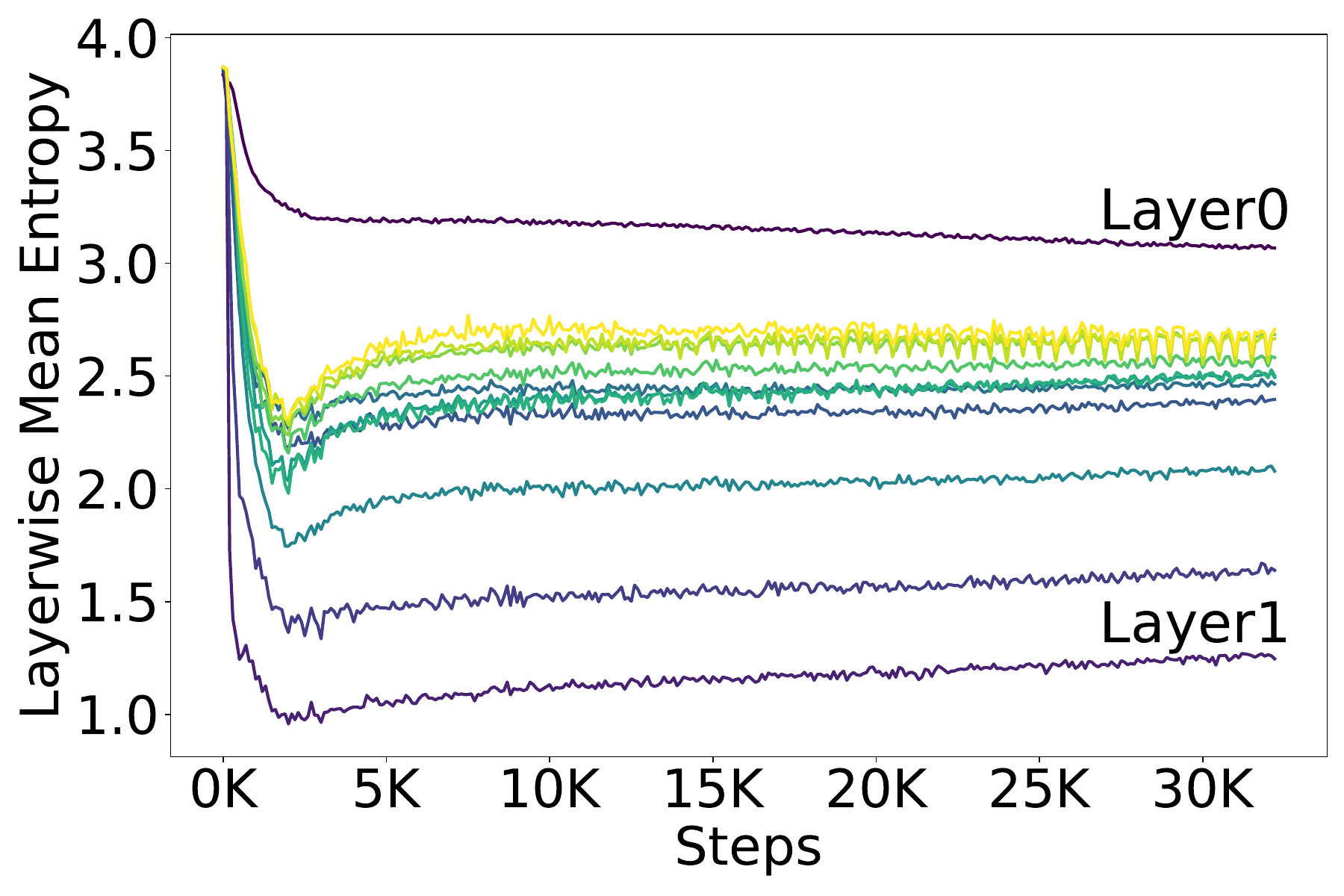}} 
\subfloat[SM \label{subfig:sm}]{\includegraphics[width=.33\textwidth]{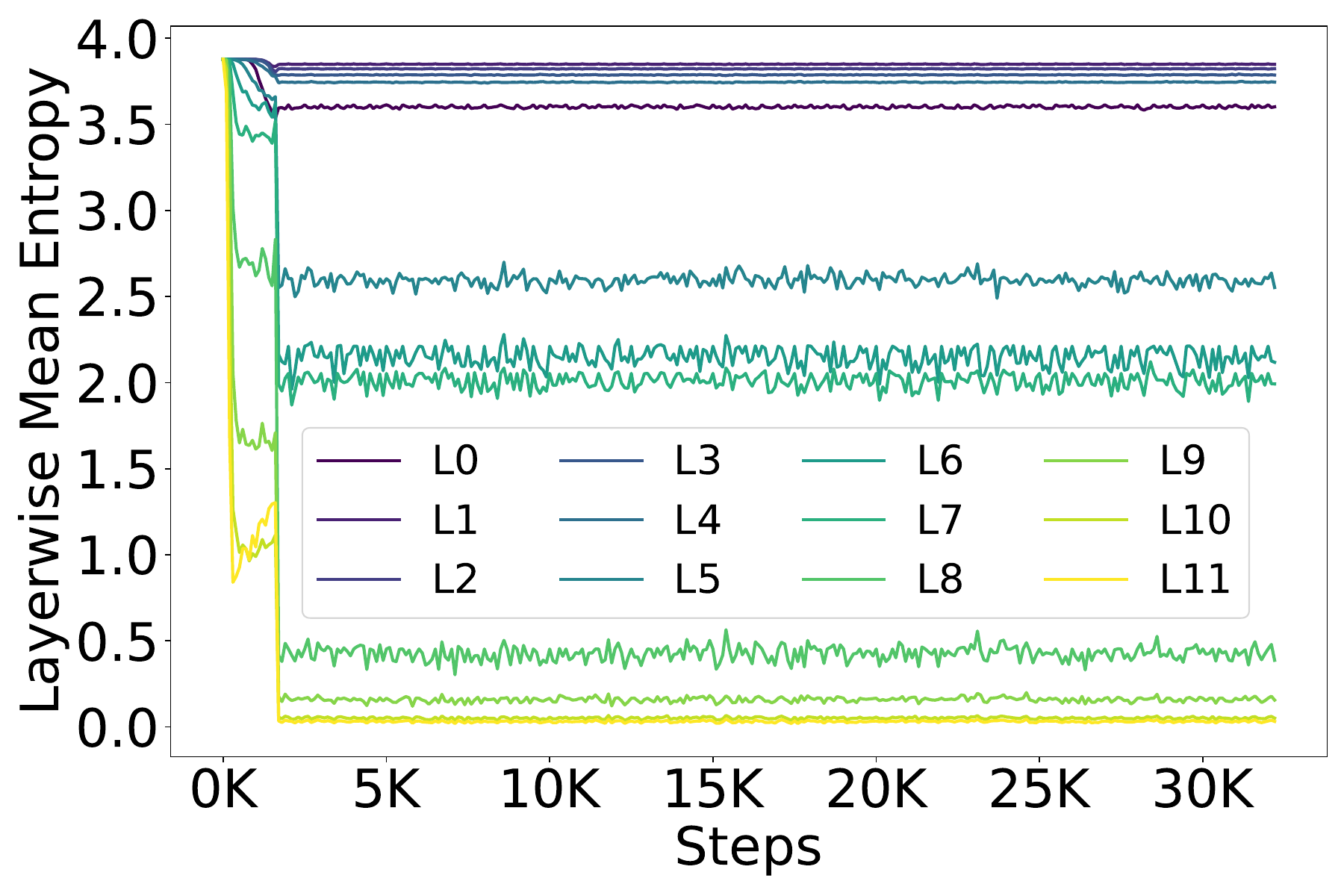}} 
\subfloat[SM + WeightNormalization(FFN) \label{subfig:sm_wnorm}]{\includegraphics[width=.33\textwidth]{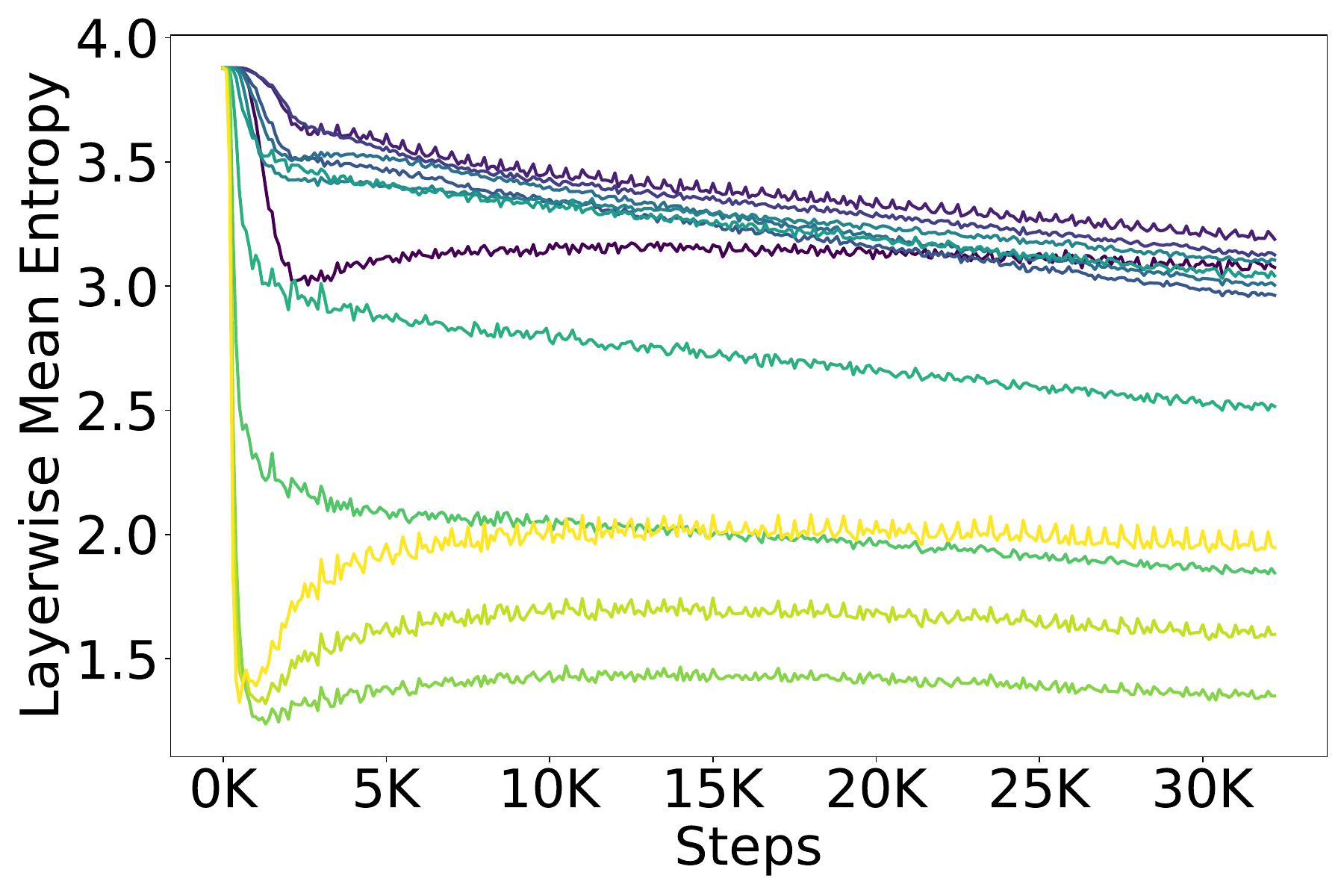}}   \\ \vspace{-1em}
\subfloat[SM + SpectralNormalization(FFN) \label{subfig:sm_snorm}]{\includegraphics[width=.33\textwidth]{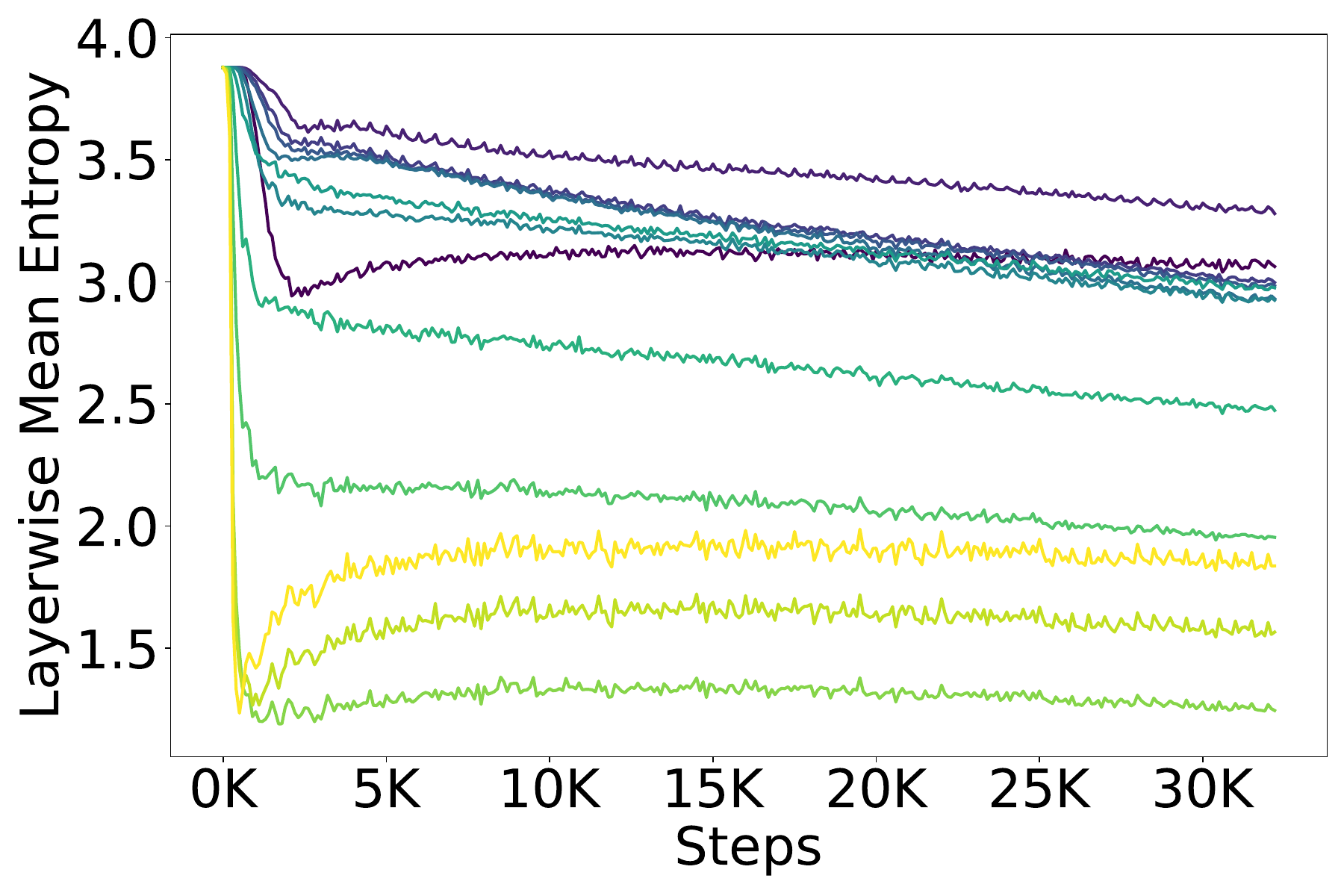}}
\subfloat[SM + Scaled(FFN) \label{subfig:sm_scaledffn}]{\includegraphics[width=.33\textwidth]{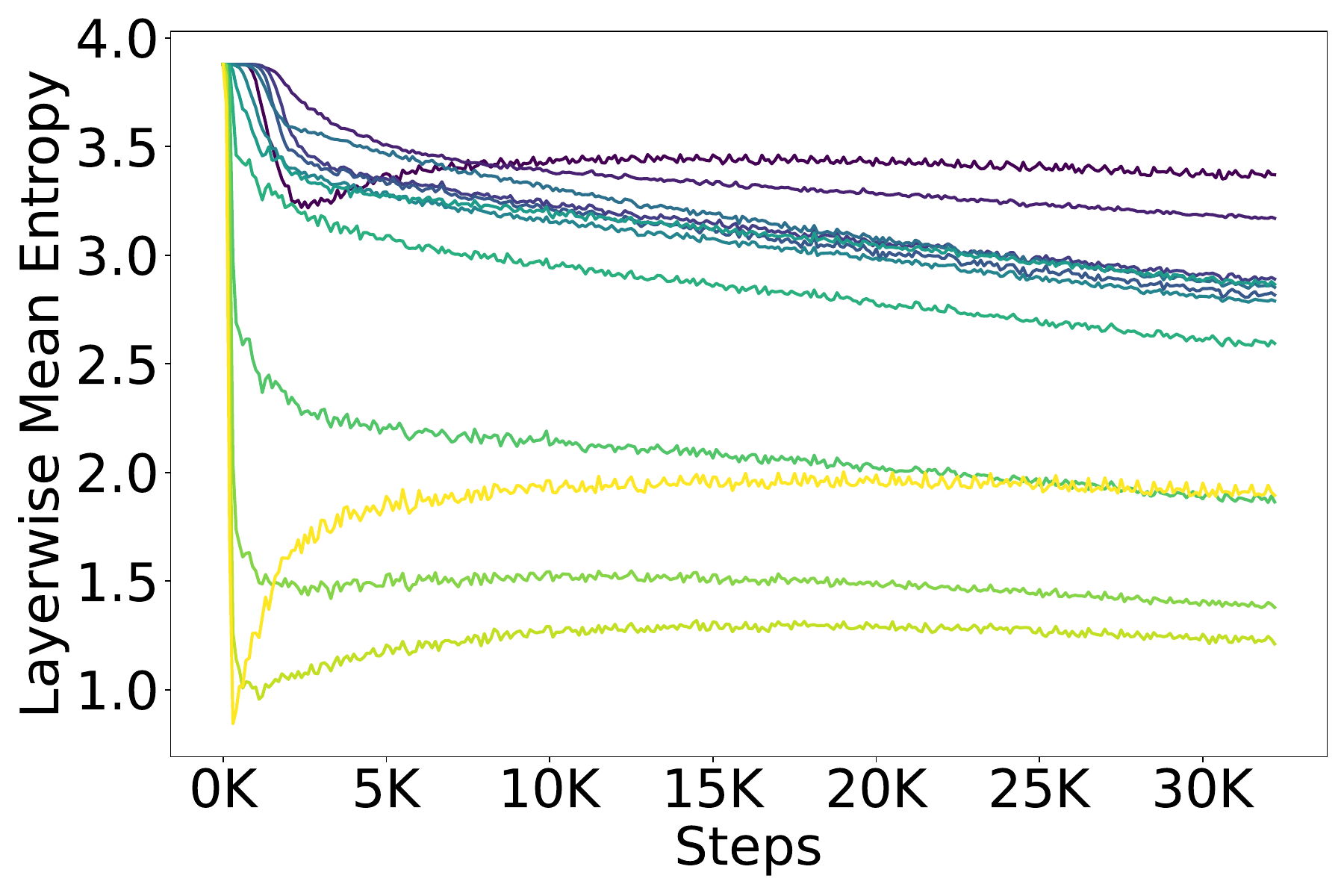}} 
\subfloat[EntropyReg($\texttt{SM(t)+ScFuFFN}$)\label{subfig:sm_ent}]{\includegraphics[width=.33\textwidth]{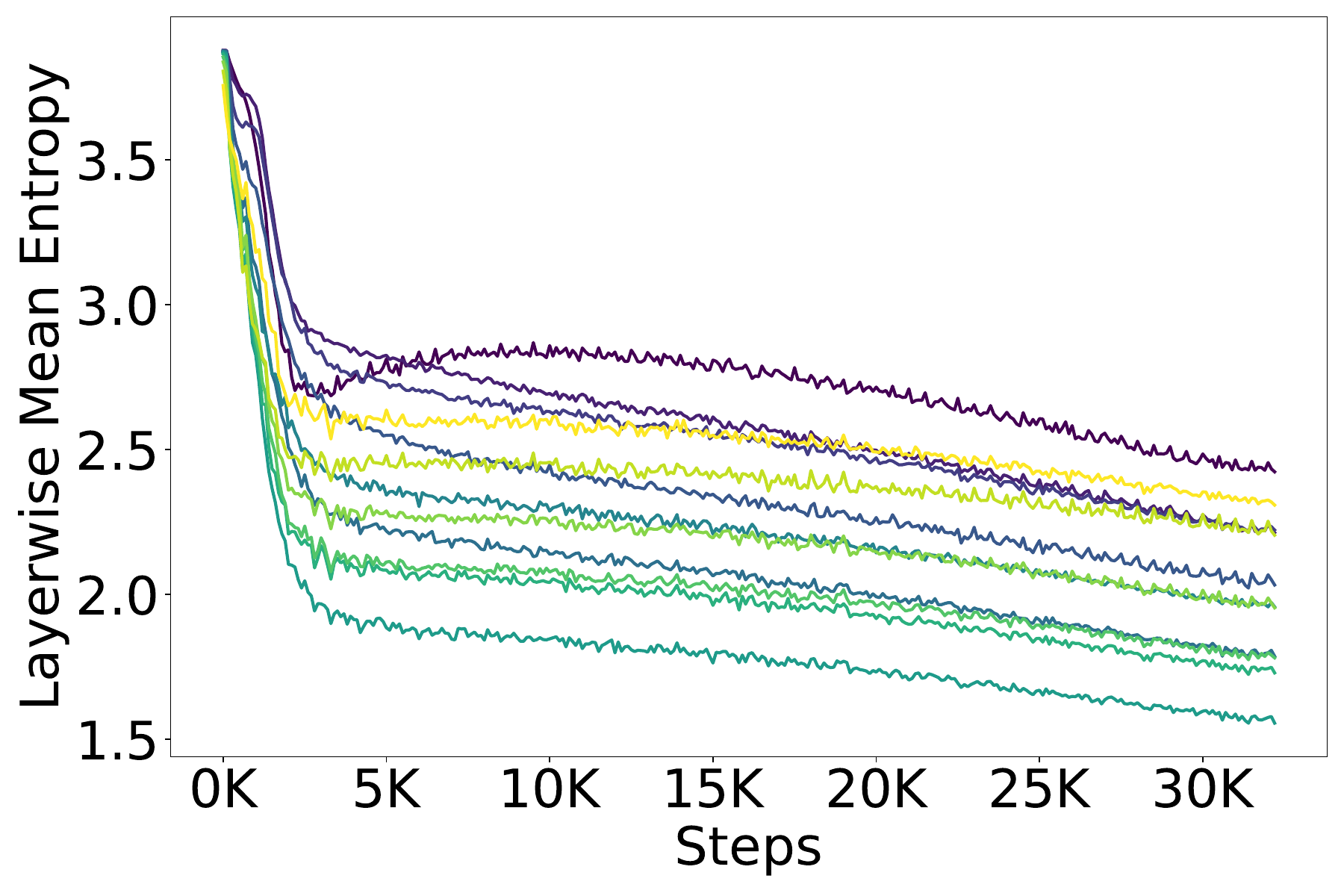}} 
\caption{Layerwise entropy patterns in GPT-2 models ($L$ = 12, $H$ = 12, $d$ = 768) trained from scratch on CodeParrot dataset. Shown are (a) baseline model, (b) Softmax-only model without normalization, and variants with (c) weight normalization, (d) spectral normalization, and (e) scaled-FFN. While these normalization methods prevent entropy collapse, they fail to address entropic overload in early layers. Our final configuration (f) incorporates entropy regularization within scaled-FFN to effectively manage both issues. } 
\label{fig:LayerwiseEntropyWSNorm}
\end{figure}

{\bf Entropy regularization prevents entropic overload in Softmax-only models} 
While both weight and spectral normalization, and scaling methods effectively prevent entropy collapse in the deeper layers and stabilize the training of Softmax-only models, they {\em fail to address the issue of entropic overload}, (see Figures \ref{subfig:sm_wnorm} and \ref{subfig:sm_snorm}). In contrast, the entropy regularization scheme penalizes the model to avoid extreme entropy values during training, resulting in a more balanced distribution. As a result, it complements the training stabilizing methods by further mitigating entropic overload in the early layers (see Figure \ref{subfig:sm_ent}), improving the utilization of attention heads and leading to improved performance, as demonstrated by lower perplexity (see Table \ref{tab:GPT2CLen128}).

\vspace{-2em}

\begin{figure} [htbp] \centering
\subfloat[Values of learned threshold weights]{\includegraphics[width=.46\textwidth]{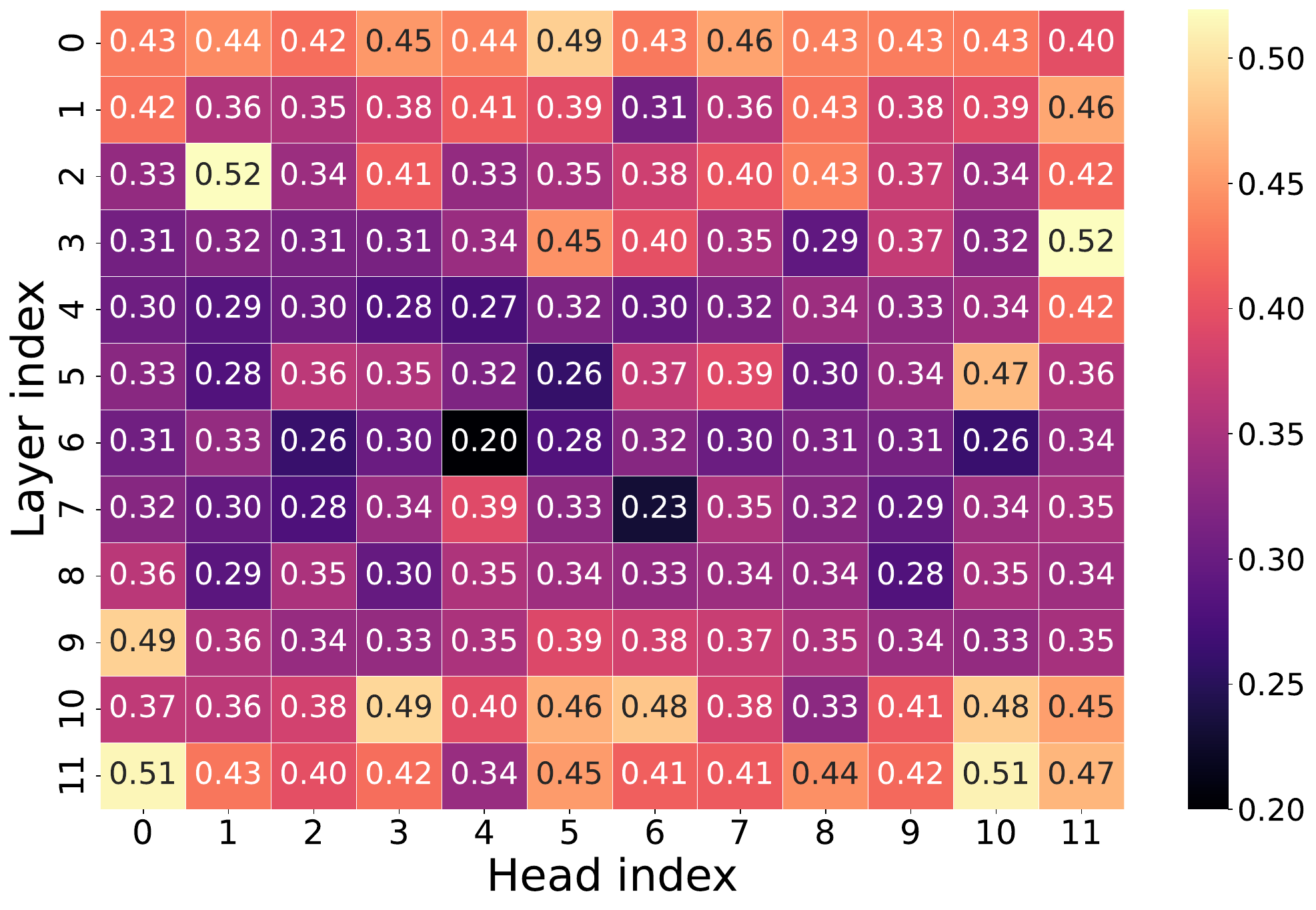}} 
\subfloat[Layerwise mean and variance  of threshold weights]{\includegraphics[width=.48\textwidth]{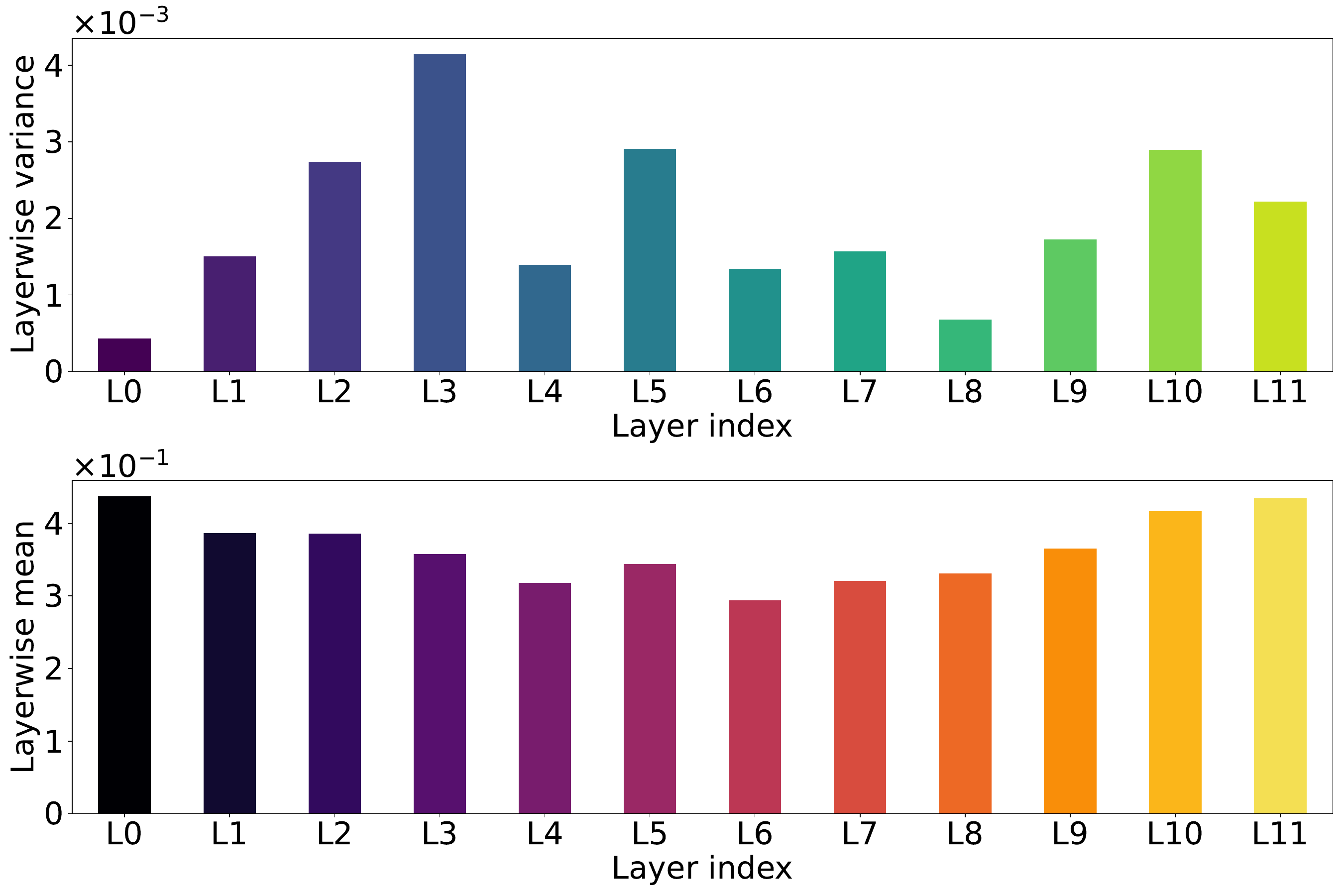}} \\ \vspace{-0.6em}
\caption{Analysis of learned threshold weights ($\mathtt{reg\_threshold\_weights}$, see Eq. \ref{eqn:EntThersDeviation}) in entropy regularization for softmax-only GPT-2 model: (a) Attention heads adaptively learn non-uniform threshold weights across different heads, setting individualized thresholds for entropy regularization; (b) The non-uniform means and non-zero variances across layers highlight the necessity and effectiveness of headwise learnable thresholds in adapting regularization strength.} 
\label{fig:EntLearnableThreshold}
\end{figure}

{\bf Significance of learnable thresholds in entropy regularization}
Figure \ref{fig:EntLearnableThreshold} depicts the learnable threshold parameters ($\mathtt{reg\_threshold\_weights}$) applied in the entropy regularization scheme after the model has been fully trained from scratch. They exhibit significant variability,  both across layers and within individual heads of each layers, which reflects the model's ability to dynamically adjust the regularization strength in response to the specific roles of different attention heads. Such flexibility is essential for tailoring the regularization process to the distinct requirements of each head.

\vspace{-1em}

{\bf Mitigating over-regularization with an appropriate threshold margin} 
\begin{wrapfigure}[15]{r}{0.45\textwidth - .95\columnsep}
\centering
\vspace{-1.5\intextsep}
\includegraphics[scale=0.24]{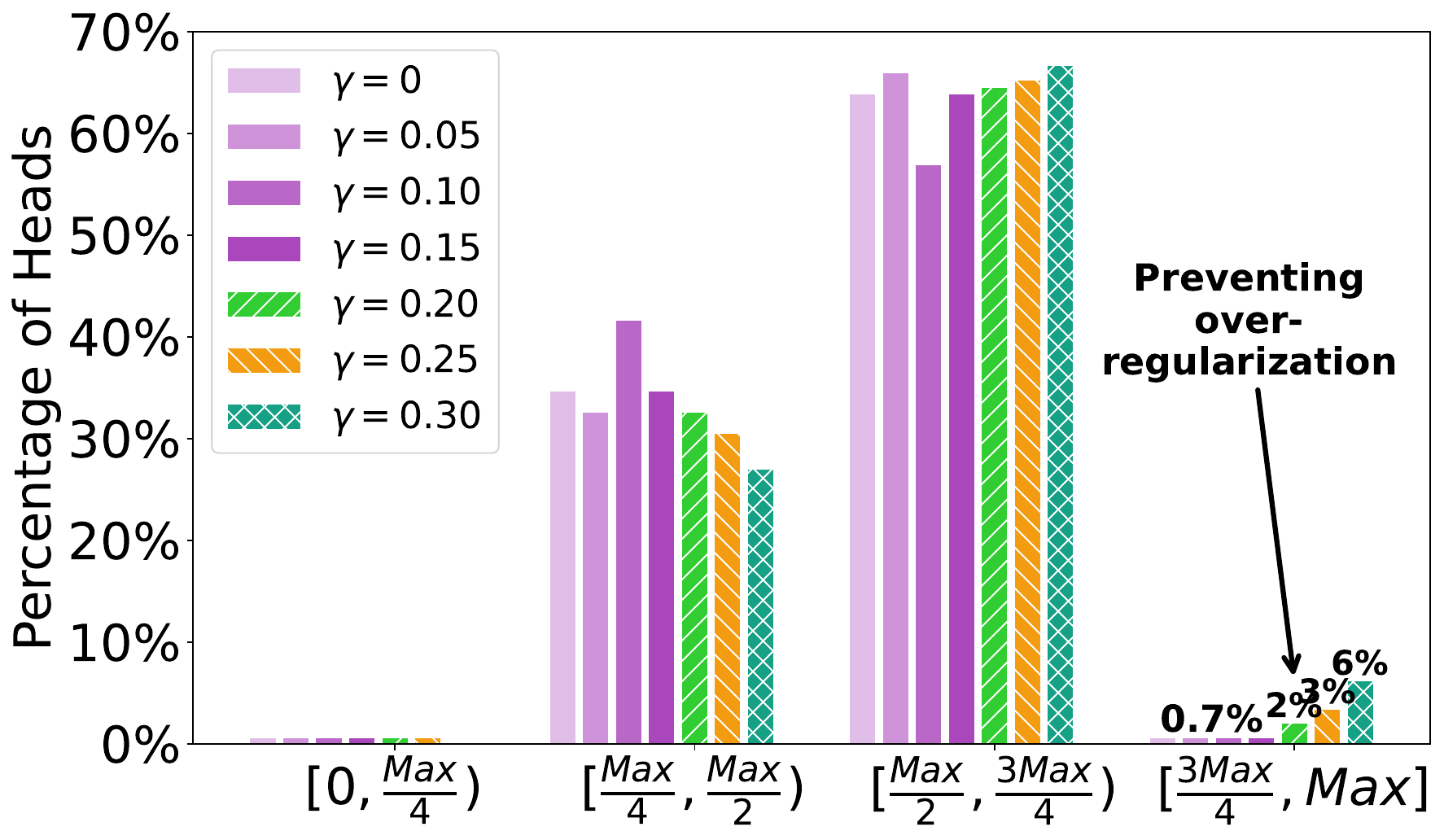} 
\vspace{-1.6em}
\caption{Headwise entropy distribution in the ${\tt SM(t) + ScFuFFN}$ GPT-2 model ($L$=12, $H$=12, $d$=768) when entropy regularization is applied with varying threshold margin, controlled by  $\gamma$.  } 
\label{fig:EntDistTmargin}
\end{wrapfigure}

Figure \ref{fig:EntDistTmargin} illustrates the effect of $\gamma$ on the headwise entropy distribution.  The hyperparameter $\gamma$  employed to adjust the threshold margin in entropy regularization, defined as \(\text{Tol}_{\text{margin}} = \gamma \text{E}_{\text{max}} \) (Algorithm\ref{Algo:EntRegLossComputation}, line \#\ref{line:ToleranceMargin}), effectively preventing over-regularization by ensuring that a sufficient fraction of heads maintains entropy values in the upper range  $\frac{\tt 3Max}{4}$ to ${\tt Max}$. As $\gamma$ increases from 0 to 0.15, only a small proportion of attention heads (0.7\%) are situated in the highest entropy range. However, as $\gamma$ is increased beyond 0.15, the fraction of heads in this upper range starts increasing, reaching 2.08\%, 3.47\%, and 6.25\% at $\gamma$=0.20, 0.25, and 0.30, respectively. This fine-grained control on the population of attention heads in the higher entropy range highlights the ability of entropy regularization to prevent over-regularization and maintain the attention heads' diversity. We find that $\gamma$=0.2  yields slightly better performance in terms of lower perplexity compared to higher $\gamma$ values, and thus, we adopt this value in the final entropy regularization scheme.

%\vspace{-1em}

\begin{figure} [htbp]
\centering
\subfloat[\(\text{Tol}_{\text{margin}} = 0 \)]{\includegraphics[width=.33\textwidth]{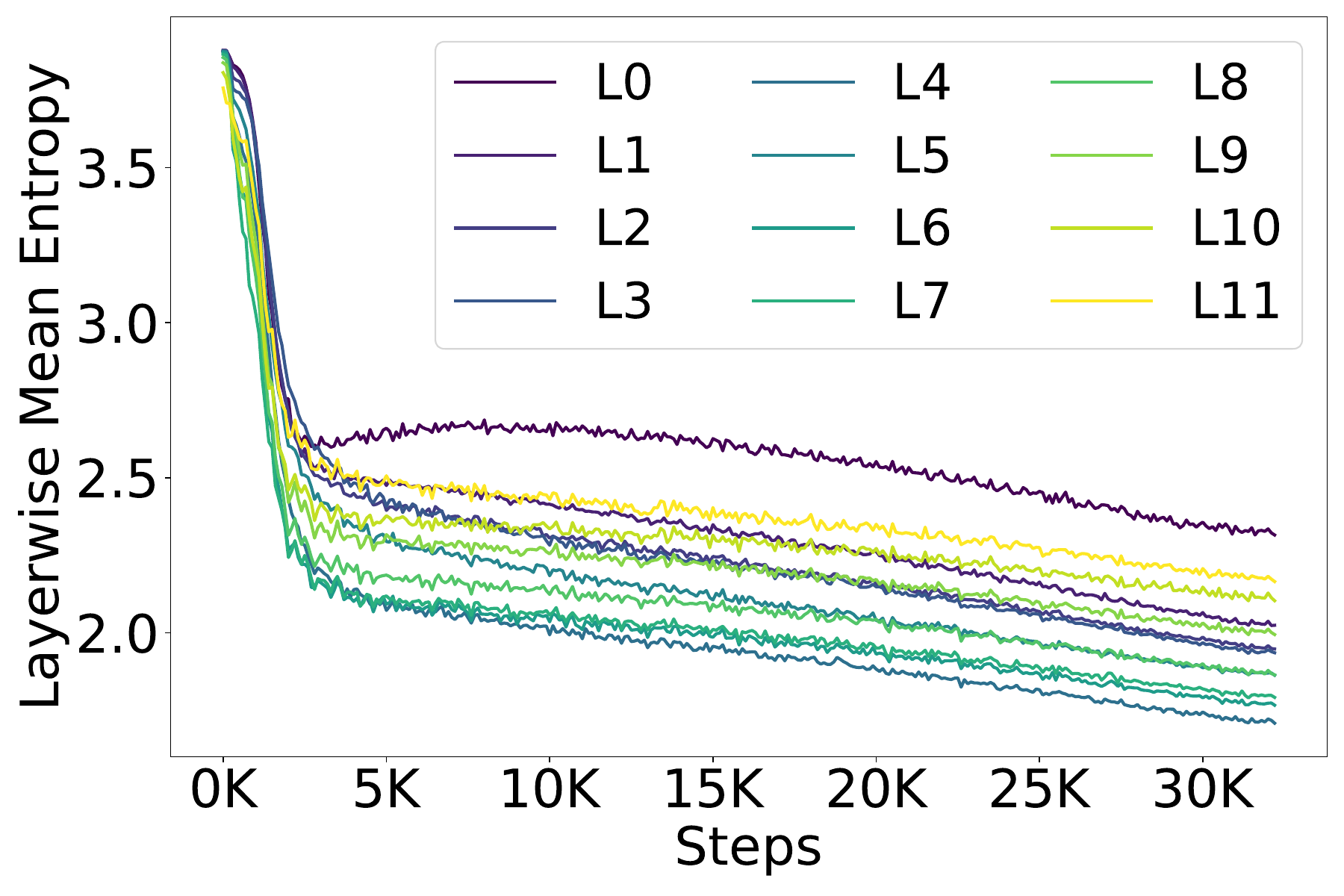}} 
\subfloat[\(\text{Tol}_{\text{margin}} = 0.05\text{E}_{\text{max}} \)]{\includegraphics[width=.33\textwidth]{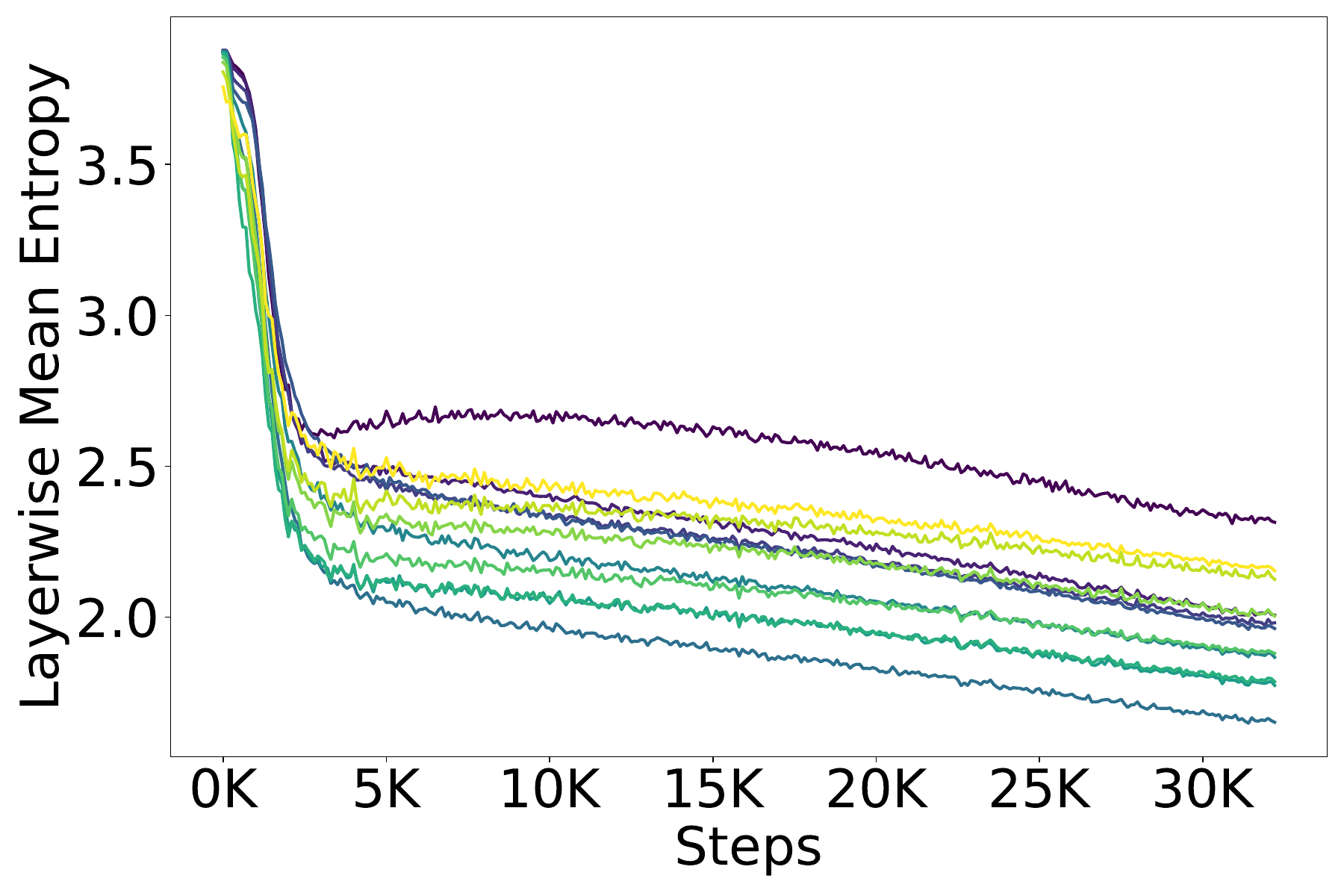}}  
\subfloat[\(\text{Tol}_{\text{margin}} = 0.10\text{E}_{\text{max}} \)]{\includegraphics[width=.33\textwidth]{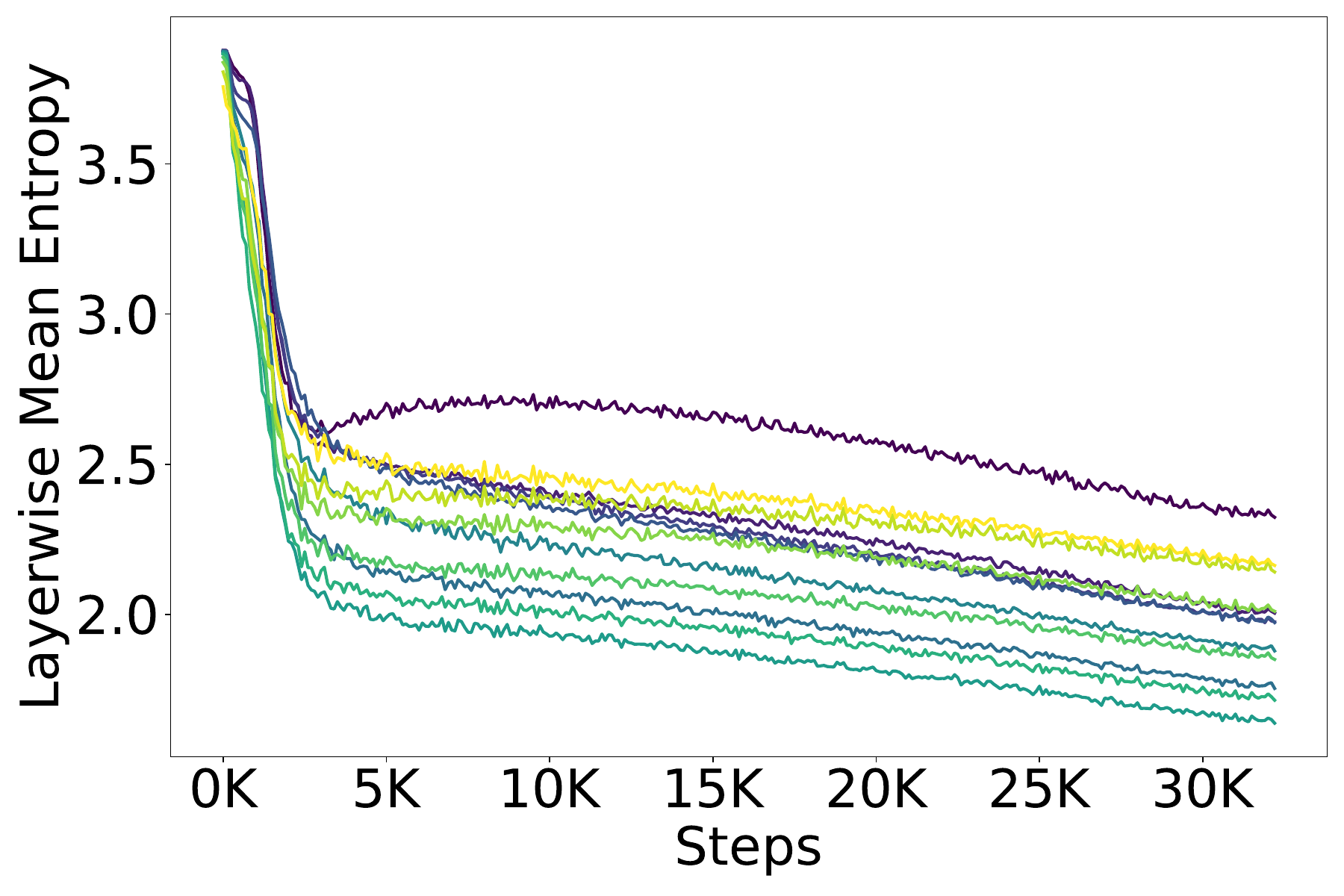}} \\ \vspace{-1em}
\subfloat[\(\text{Tol}_{\text{margin}} = 0.15\text{E}_{\text{max}} \)]{\includegraphics[width=.33\textwidth]{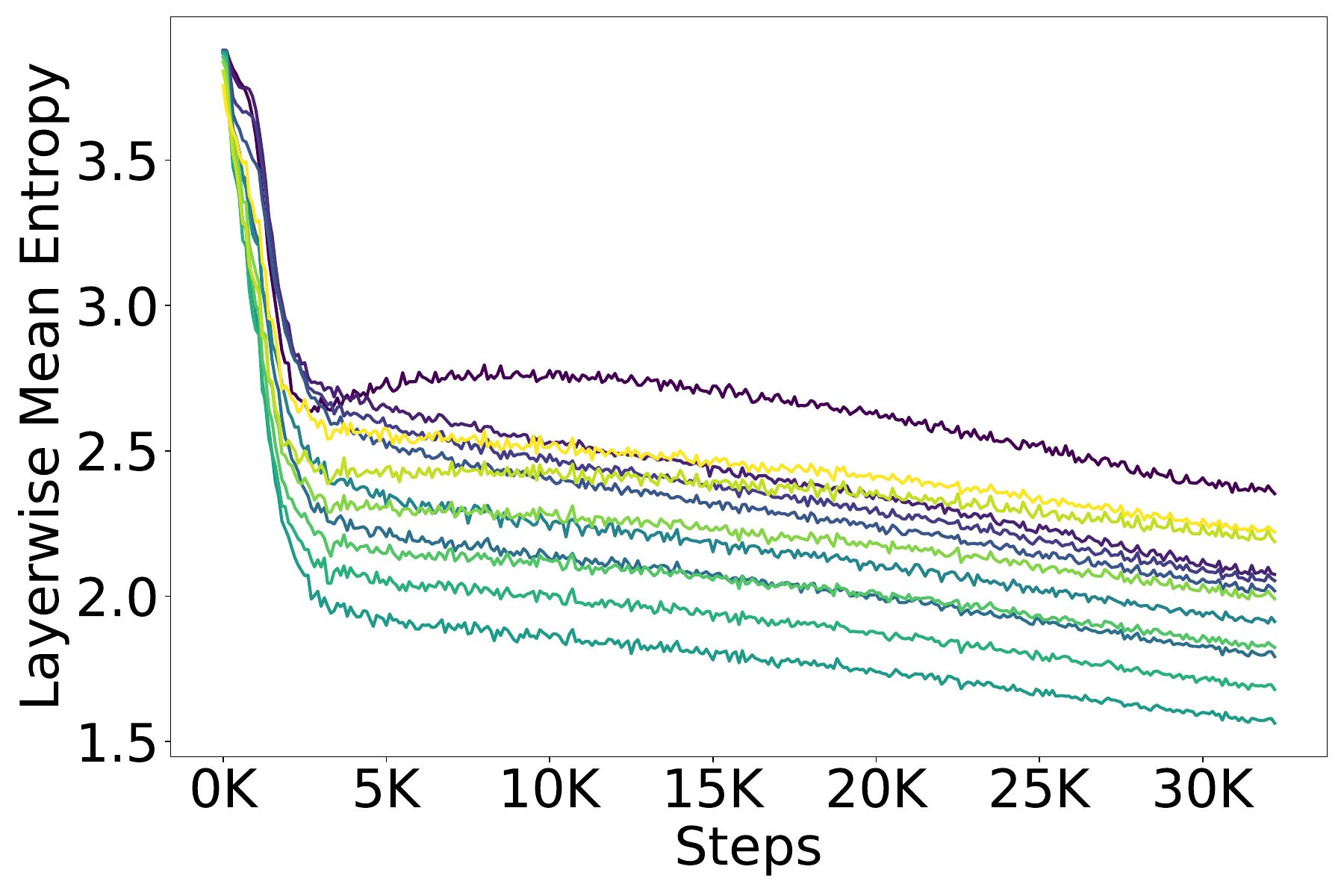}} 
\subfloat[\(\text{Tol}_{\text{margin}} = 0.20\text{E}_{\text{max}} \)]{\includegraphics[width=.33\textwidth]{plots/ent_layerwise_Tmargin_20Per}}  
\subfloat[\(\text{Tol}_{\text{margin}} = 0.25\text{E}_{\text{max}} \)]{\includegraphics[width=.33\textwidth]{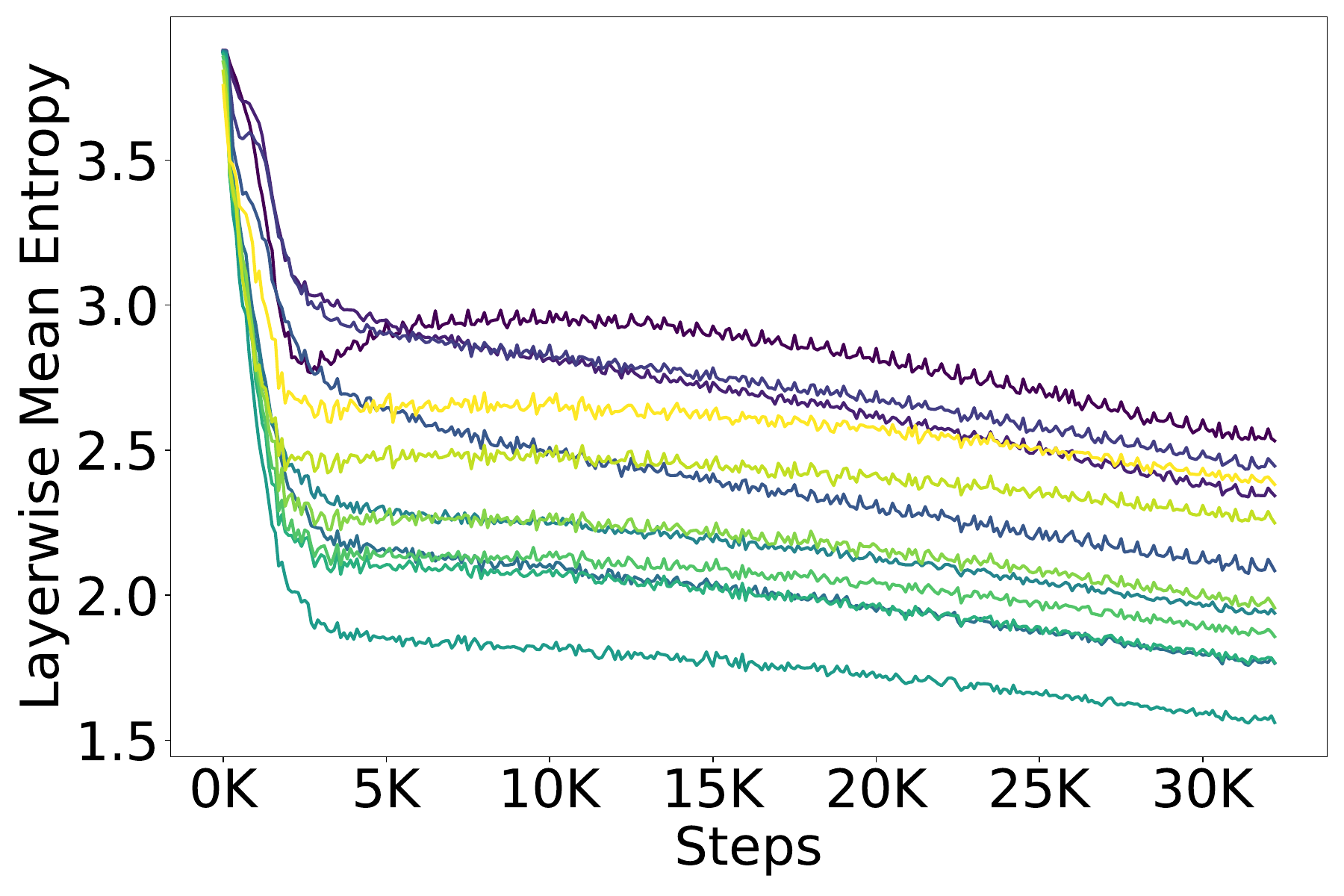}}  
%\subfloat[\(\text{Tol}_{\text{margin}} = 0.30\text{E}_{\text{max}} \)]{\includegraphics[width=.25\textwidth]{plots/ent_layerwise_Tmargin_30Per}}  \vspace{-0.6em}
\caption{Layerwise entropy dynamics when entropy regularization is employed with increasing threshold margin, defined as \(\text{Tol}_{\text{margin}} = \gamma \text{E}_{\text{max}} \) (see Algorithm\ref{Algo:EntRegLossComputation}, line \#\ref{line:ToleranceMargin}). At higher $\gamma$, the mean entropy of the early layers increases.}
\label{fig:EntropyDynamicsTmargins}
\end{figure}

%\vspace{-1em}

To gain deeper insights, Figure \ref{fig:EntropyDynamicsTmargins} illustrates entropy dynamics  with increasing $\gamma$ during training. 
As $\gamma$ increases, the proportion of attention heads exhibiting higher entropy values grows. This is reflected in the rising mean entropy of the early layers, which plays a crucial role in preventing over-regularization and preserving the diversity of attention heads.

{\bf Results on GPT-2 model}
Table \ref{tab:GPT2CLen128} presents results for GPT-2 small models, offering a detailed breakdown of nonlinear operations and FLOPs. The architectural simplification through nonlinearity reduction (${\tt SM(t) + ScFuFFN}$) achieves a 3.94$\times$ reduction in communication overhead and a 1.72$\times$ speedup in end-to-end PI latency. Additionally, entropy regularization enhances the perplexity of the ${\tt SM(t) + ScFuFFN}$ model by {\bf 7.8\%}, validating the effectiveness of the entropy-guided attention mechanism.

\begin{table}[htbp]
\centering
\caption{Results on GPT-2 ($L$=12, $H$=12, $d$=768), trained from scratch on the CodeParrot dataset (2.1B tokens, $T$=128).} 
\label{tab:GPT2CLen128}
\resizebox{0.99\textwidth}{!}{
\begin{tabular}{p{0.2cm}lclcccccc}
\toprule
& \multirow{2}[4]{*}{Network Arch.} & \multirow{2}[4]{*}{PPL} & \multirow{2}[4]{*}{\#Nonlinear Ops} & \multicolumn{2}{c}{\#FLOPs} & \multirow{2}[4]{*}{\shortstack{Comm.\\(GB)}} & \multirow{2}[4]{*}{\shortstack{Lat.\\(min.)}} & \multicolumn{2}{c}{Savings} \\
\cmidrule(lr){5-6} \cmidrule(lr){9-10}
& & & & FFN & Attn. & & & Comm. & Lat. \\
\midrule
\multirow{6}{*}{  \rotatebox[origin=c]{90}{Baseline} } & \multirow{3}{*}{${\tt SM + LN + G}$} & \multirow{3}{*}{2.69} & SM:$144\times\mathbb{R}^{128\times128}$ & \multirow{3}{*}{14.5B} & \multirow{3}{*}{7.7B} & \multirow{3}{*}{25.32} & \multirow{3}{*}{8.21} & \multirow{3}{*}{1$\times$} & \multirow{3}{*}{1$\times$} \\
& & & LN:$24\times\mathbb{R}^{128\times768}$ & & & & & & \\
& & & G:$12\times\mathbb{R}^{128\times3072}$ & & & & & & \\ \cline{2-10}
& \multirow{3}{*}{${\tt SM + LN + R}$} & \multirow{3}{*}{2.76} & SM:$144\times\mathbb{R}^{128\times128}$ & \multirow{3}{*}{14.5B} & \multirow{3}{*}{7.7B} & \multirow{3}{*}{9.44} & \multirow{3}{*}{6.06} & \multirow{3}{*}{2.68$\times$} & \multirow{3}{*}{1.35$\times$} \\
& & & LN:$24\times\mathbb{R}^{128\times768}$ & & & & & & \\
& & & R:$12\times\mathbb{R}^{128\times3072}$ & & & & & & \\
\midrule
&${\tt SM + ScFuFFN}$ & 3.48 &  SM:$144\times\mathbb{R}^{128\times128}$ & 1.8B & 7.7B & 6.43 & 4.76  & 3.94$\times$  & 1.72$\times$ \\
\rowcolor{green!20} &$\text{EReg}({\tt SM(t) + ScFuFFN})$ & 3.21 &  SM:$144\times\mathbb{R}^{128\times128}$ & 1.8B & 7.7B  & 6.43 & 4.76  & 3.94$\times$ & 1.72$\times$ \\
\bottomrule
\end{tabular}}
\end{table}

{\bf Scalability across model depth, context length, and training data}
To demonstrate the robustness of our approach, we evaluate both architectural simplifications and entropy-guided solutions across different model configurations. Experiments with deeper models (Table \ref{tab:GPT218LCLen128}) and increased context lengths (Tables~\ref{tab:GPT2CLen256}) show consistent benefits in terms of nonlinearity reduction and entropy regularization effectiveness.

We further analyze the scalability of our approach across different training regimes using the Languini dataset. Table \ref{tab:LanguiniGPT2} presents latency and communication improvements for GPT-2 models trained on varying token counts (1.2B, 2.4B, and 4.8B), demonstrating the consistency of our architectural benefits across different training scales.

\begin{table}[htbp]
\centering
%Results on GPT-2 ($L$=12, $H$=12, $d$=768), trained from scratch on the CodeParrot dataset ($T$=256)
\caption{Results on  GPT-2 ($L$=12, $H$=12, $d$=768) model, trained from scratch on Languini \cite{stanic2023languini} ($T$=512)}
\label{tab:LanguiniGPT2}
\resizebox{0.99\textwidth}{!}{
\begin{tabular}{p{0.2cm}lccclcccc}
\toprule
& \multirow{2}[4]{*}{Network Arch.} & \multicolumn{3}{c}{Eval PPL} & \multirow{2}[4]{*}{\#Nonlinear Ops} & \multicolumn{2}{c}{\#FLOPs} & \multirow{2}[4]{*}{\shortstack{Comm.\\(GB)}} & \multirow{2}[4]{*}{\shortstack{Lat.\\(min.)}} \\
\cmidrule(lr){3-5} \cmidrule(lr){7-8}
& & 1.2B & 2.4B & 4.8B & & FFN & Attn. & & \\
\midrule
\multirow{6}{*}{\rotatebox[origin=c]{90}{Baseline}} & \multirow{3}{*}{${\tt SM + LN + G}$} & \multirow{3}{*}{25.71} & \multirow{3}{*}{23.32} & \multirow{3}{*}{21.29} & SM:$144\times\mathbb{R}^{512\times512}$ & \multirow{3}{*}{58.0B} & \multirow{3}{*}{36.2B} & \multirow{3}{*}{145.24} & \multirow{3}{*}{30.74} \\
& & & & & LN:$24\times\mathbb{R}^{512\times768}$ & & & & \\
& & & & & G:$12\times\mathbb{R}^{512\times3072}$ & & & & \\ \cline{2-10}
& \multirow{3}{*}{${\tt SM + LN + R}$} & \multirow{3}{*}{26.06} & \multirow{3}{*}{23.55} & \multirow{3}{*}{21.58} & SM:$144\times\mathbb{R}^{512\times512}$ & \multirow{3}{*}{58.0B} & \multirow{3}{*}{36.2B} & \multirow{3}{*}{81.71} & \multirow{3}{*}{23.54} \\
& & & & & LN:$24\times\mathbb{R}^{512\times768}$ & & & & \\
& & & & & R:$12\times\mathbb{R}^{512\times3072}$ & & & & \\
\midrule
&${\tt SM + ScFuFFN}$ & 33.77 & 30.82 & 28.59 & SM:$144\times\mathbb{R}^{512\times512}$ & 7.3B & 36.2B & 69.68 & 19.44 \\
%&${\tt SM + ScFuFFNi_1}$ & 34.16 & 31.23 & 29.69 & SM:$144\times\mathbb{R}^{512\times512}$ & 6.6B & 36.2B  & 69.64 & 19.11 \\ \cline{2-10}
\rowcolor{green!20} &$\text{EReg}({\tt SM(t) + ScFuFFN})$ & 31.54 & 28.70 & 26.55 & SM:$144\times\mathbb{R}^{512\times512}$ & 7.3B & 36.2B & 69.68 & 19.44 \\
%&$\text{EReg}({\tt SM(t) + ScFuFFNi_1})$ & 31.75 & 28.93 & 26.74 & SM:$144\times\mathbb{R}^{512\times512}$ & 6.6B & 36.2B  & 69.64 & 19.11  \\
\bottomrule
\end{tabular}}
\end{table}

\section{Conclusion}
In this work, we address the fundamental challenges posed by nonlinear operations in private LLMs inference. By leveraging an information-theoretic framework, we uncover the dual role of nonlinearities in ensuring training stability and maintaining attention head diversity. Our study introduces novel entropy regularization techniques, and PI-friendly alternatives for layer normalization, demonstrating their effectiveness in mitigating entropy collapse and entropic overload. These contributions pave the way for PI-optimized architectures with reduced-nonlinearities, significantly reducing latency and communication overheads. By addressing the critical trade-offs between nonlinearity, computational overhead, and entropy dynamics, we provide a clear path toward scalable and practical PI systems.

{\bf Limitations}
This study mainly focuses on pre-training performance, with perplexity as the primary metric, and does not include experiments to evaluate other capabilities such as transfer learning or few-shot learning. Additionally, the efficacy of the proposed Softmax-only models has been validated on LLMs with lesser than 1B parameters. Future work will explore broader experimental evaluations,  including their adaption for large-scale models.

\section*{Additional Notes}
This workshop version focuses on the fundamental role of nonlinearities in maintaining model stability and fostering attention head diversity in LLMs, as well as their implications for private inference. These findings are part of a broader study presented in our comprehensive paper  \href{https://arxiv.org/abs/2410.13060}{AERO: Softmax-Only LLMs for Efficient Private Inference}. The code and implementation are available at \href{https://github.com/Nandan91/entropy-guided-attention-llm}{entropy-guided-llm}.

\bibliography{MyRef}
\bibliographystyle{unsrt}

\newpage
\appendix

\renewcommand{\partname}{}

\addcontentsline{toc}{section}{Appendix} % Add the appendix text to the document TOC
\part{Appendix} % Start the appendix part
\parttoc % Insert the appendix TOC

\newpage

\section{Softmax Learnable Temperature for Entropy-Guided Attention} \label{subsecAppendix:EntRegFullDerivation}

With the learnable temperature parameters ($t$), the attention matrix can be expressed as follows:
 
\vspace{-1.2em}

\begin{equation}
\mathbf{A}^{(l, h)}(t) = \left[ a_{ij}^{(l, h)}(t) \right]_{T \times T}, \; \text{where} \; a_{ij}^{(l, h)}(t) = \frac{\exp\left(\frac{1}{t_i \sqrt{d_k}} (\mathbf{X}_i \mathbf{W}^Q) (\mathbf{X}_j \mathbf{W}^K)^\top \right)}{\sum_{k=1}^{T} \exp\left(\frac{1}{t_i \sqrt{d_k}} (\mathbf{X}_i \mathbf{W}^Q) (\mathbf{X}_k \mathbf{W}^K)^\top \right).} 
\end{equation}

Let $z_{ij} = \left( \mathbf{X}_i \mathbf{W}^Q \right) \left( \mathbf{X}_j \mathbf{W}^K \right)^\top$ represents the logits (attention scores before applying softmax).

Now, substituting \( a_{ij}^{(l,h)}(t) \) into the entropy formula:
\vspace{-1em}
\begin{equation}
\mathbf{E}^{(l,h)}(t) = -\frac{1}{T} \sum_{i=1}^{T} \sum_{j=1}^{T} \frac{\exp\left(\frac{1}{t \sqrt{d_k}} z_{ij}\right)}{\sum_{k=1}^{T} \exp\left(\frac{1}{t \sqrt{d_k}} z_{ik}\right)} \log \left( \frac{\exp\left(\frac{1}{t \sqrt{d_k}} z_{ij}\right)}{\sum_{k=1}^{T} \exp\left(\frac{1}{t \sqrt{d_k}} z_{ik}\right)} \right). \nonumber
\end{equation}

Simplifying the logarithmic term:

\begin{equation}
\log \left( \frac{\exp\left(\frac{1}{t \sqrt{d_k}} z_{ij}\right)}{\sum_{k=1}^{T} \exp\left(\frac{1}{t \sqrt{d_k}} z_{ik}\right)} \right) = \frac{1}{t \sqrt{d_k}} z_{ij} - \log \left( \sum_{k=1}^{T} \exp\left(\frac{1}{t \sqrt{d_k}} z_{ik}\right) \right). \nonumber
\end{equation}

Thus, the entropy simplifies to:

\begin{equation}
\mathbf{E}^{(l,h)}(t) = \frac{1}{T} \sum_{i=1}^{T} \left( \log \left( \sum_{k=1}^{T} \exp\left(\frac{1}{t \sqrt{d_k}} z_{ik}\right) \right) - \frac{1}{t \sqrt{d_k}} \sum_{j=1}^{T} a_{ij}^{(l,h)}(t) z_{ij} \right).  \nonumber
\end{equation}

Further, it can be simplified as a function of expected value of $z_{ij}$ under the attention distribution:

\begin{equation} \label{eqn:SimplifiedEnt}
\mathbf{E}^{(l,h)}(t) = \frac{1}{T} \sum_{i=1}^{T} \left( \log \left( \sum_{k=1}^{T} \exp\left( \frac{z_{ik}}{t \sqrt{d_k}} \right) \right) - \frac{1}{t \sqrt{d_k}} \, \mathbb{E}_{j \sim a_{ij}^{(l,h)}(t)} [\, z_{ij} \,] \right)
\end{equation}

In the above expression (Eq. \ref{eqn:SimplifiedEnt}), the first term \( \left( \log \sum \right) \) represents the overall {\em spread} of the logits when scaled by \(t\), and the second term \( \left( \frac{1}{t} \mathbb{E}[z_{ij}] \right) \) represents the expected value of the scaled logits under the attention distribution. 

{\bf Temperature cases when:} 

\begin{enumerate} 
\item \(t > 1\): The scaling factor \( \frac{1}{t} \) reduces the influence of the logits \(z_{ij}\), making the softmax distribution more uniform. Consequently, the entropy {\em increases}.
\item \(t < 1\): The scaling factor \( \frac{1}{t} \) increases the influence of the logits \(z_{ij}\), making the softmax distribution more peaked. Consequently, the entropy {\em decreases}.
\item  \(t \to \infty\): The logits are scaled down to zero, and the softmax becomes a uniform distribution. The entropy reaches its maximum value of \( \log T \).
\item \(t \to 0\): The logits dominate the softmax, and it becomes a one-hot distribution. The entropy approaches zero.
\end{enumerate}

%[noitemsep,nolistsep,leftmargin=0.5cm]

\newpage
\section{PyTorch Implementation of Entropy Regularization} \label{subsecAppendix:EntRegPythonCode}

\lstset{
  basicstyle=\ttfamily\small,
  columns=fullflexible,
  breaklines=true,
  numbers=left,
  numberstyle=\tiny\color{gray},
  keywordstyle=\color{blue},
  commentstyle=\color{teal},
  stringstyle=\color{red},
  showstringspaces=false,
  frame=single,
  language=Python
}

The PyTorch implementation below computes the entropy regularization loss for attention weights in a transformer model. This regularization ensures a balanced attention score distribution, fostering head-specialization in MHA.

\begin{lstlisting}[language=Python, caption={Entropy Regularization Loss Calculation}, label={lst:entropy_reg_loss}]
import torch

def calculate_entropy_reg_loss(attentions, blocks, seq_len):
    """
    Calculate the entropy regularization loss.

    Parameters:
    attentions (list): A list of attention matrices from different layers.
    blocks (list): A list of transformer blocks.
    seq_len (int): The length of the sequence (context length).

    Returns:
    float: The entropy regularization loss.
    """
    entropy_reg_loss = 0
    max_entropy = torch.log(torch.tensor(seq_len))  # Theoretical maximum entropy 
    fraction = 0.10  # Design hyper-parameter for tolerance margin
    tolerance_margin = fraction * max_entropy  # Set tolerance margin as fraction of the maximum entropy

    for layer_idx, (block, attn_mat) in enumerate(zip(blocks, attentions)):
        reg_threshold_weights = block.attn.reg_threshold_weights # Head-wise learnable parameters to set head-specific threshold
        ent_val = -torch.sum(attn_mat * torch.log(attn_mat + 1e-9), dim=-1)  # Compute entropy averaged over sequence length
        layer_entropy_reg_loss = 0

        for head_idx in range(block.attn.num_heads):
            head_entropy = ent_val[:, head_idx, :]  # Get head-specific entropy
            threshold = reg_threshold_weights[head_idx] * max_entropy
            deviation = torch.abs(head_entropy - threshold)
            penalty = torch.square(torch.where(deviation > tolerance_margin, deviation, torch.zeros_like(deviation)))
            layer_entropy_reg_loss += penalty.sum()

        layer_entropy_reg_loss /= block.attn.num_heads
        entropy_reg_loss += layer_entropy_reg_loss

    entropy_reg_loss /= len(attentions)
    return entropy_reg_loss
    
# Calculate the total loss including entropy regularization
lambda_reg = 1e-5  # Hyperparameter for entropy regularization weight
entropy_regularization = calculate_entropy_reg_loss(attentions, blocks, seq_len) 
total_loss = ce_loss + lambda_reg * entropy_regularization
\end{lstlisting}

\section{Additional Results}

\subsection{Performance Comparison of  Weight and Spectral Normalization, and Learnable FFN Scaling} 
Table \ref{tab:SNormVsWNorm} compares the performance of weight and spectral normalization applied in various linear layers within the attention and FFN sub-blocks in Softmax-only model. The results show that applying these techniques to the attention blocks yields diminishing returns compared to their application in the FFN.

\begin{table}[htbp]
\caption{Comparison of weight normalization \cite{salimans2016weight} and spectral normalization \cite{miyato2018spectral} when employed in Softmax-only GPT-2 ($L$=12, $H$=12, $d$=768) models, and trained from scratch on CodeParrot dataset with 128 input context length. FFN weight normalization yield the similar results; whereas, weight normalization works better in other linear layers.}
\label{tab:SNormVsWNorm}
\centering 
\begin{tabular}{lcc} \toprule 
Linear layers & Eval PPL(Weight Normalization) & Eval PPL(Spectral Normalization) \\ \toprule 
QK & 3.89 & 4.25 \\
\rowcolor{green!20} FFN &  3.64 & 3.63 \\
QK+FFN& 3.88 & 4.23 \\
QKV+FFN& 3.93 & 4.26 \\
QKVO+FFN& 3.98 & 4.34 \\ \bottomrule  
\end{tabular} 
\end{table}

When comparing performance, we find that weight and spectral normalization led to similar performance while the learnable scaling method outperformed them with a lower perplexity (Table \ref{tab:SNormVsWNormVsMlpGains}).

\begin{table}[htbp]
\caption{Perplexity comparison  of weight normalization, spectral normalization, and learnable scaling employed in FFN of softmax-only GPT-2 model, when trained from scratch on CodeParrot dataset with 128 input context length.} 
\label{tab:SNormVsWNormVsMlpGains}
\centering
\begin{tabular}{cccc} \toprule 
& Weight Normalization & Spectral Normalization & Scaled-FFN\\ \toprule 
Eval PPL & 3.640 & 3.624 & 3.478 \\ \bottomrule 
\end{tabular} 
\end{table}

\subsection{Scalability for model depth and context length}

{\bf GPT-2 Model with 256 tokens as input context}
Table \ref{tab:GPT2CLen256} provides the latency and communication savings achieved on the GPT-2 model with 256 context length, along with a detailed breakdown of the nonlinear operations and FLOPs.

\begin{table}[htbp]
\centering
\caption{Results on GPT-2 ($L$=12, $H$=12, $d$=768), trained from scratch on the CodeParrot dataset (2.1B tokens, $T$=256).}
\label{tab:GPT2CLen256}
\resizebox{0.99\textwidth}{!}{
\begin{tabular}{p{0.2cm}lclcccccc}
\toprule
& \multirow{2}[4]{*}{Network Arch.} & \multirow{2}[4]{*}{PPL} & \multirow{2}[4]{*}{\#Nonlinear Ops} & \multicolumn{2}{c}{\#FLOPs} & \multirow{2}[4]{*}{\shortstack{Comm.\\(GB)}} & \multirow{2}[4]{*}{\shortstack{Lat.\\(min.)}} & \multicolumn{2}{c}{Savings} \\
\cmidrule(lr){5-6} \cmidrule(lr){9-10}
& & & & FFN & Attn. & & & Comm. & Lat. \\
\midrule
\multirow{6}{*}{\rotatebox[origin=c]{90}{Baseline}} & \multirow{3}{*}{${\tt SM + LN + G}$} & \multirow{3}{*}{2.35} & SM:$144\times\mathbb{R}^{256\times256}$ & \multirow{3}{*}{29.0B} & \multirow{3}{*}{16.3B} & \multirow{3}{*}{58.51} & \multirow{3}{*}{16.57} & \multirow{3}{*}{1$\times$} & \multirow{3}{*}{1$\times$} \\
& & & LN:$24\times\mathbb{R}^{256\times768}$ & & & & & & \\
& & & G:$12\times\mathbb{R}^{256\times3072}$ & & & & & & \\ \cline{2-10}
& \multirow{3}{*}{${\tt SM + LN + R}$} & \multirow{3}{*}{2.41} & SM:$144\times\mathbb{R}^{256\times256}$ & \multirow{3}{*}{29.0B} & \multirow{3}{*}{16.3B} & \multirow{3}{*}{26.73} & \multirow{3}{*}{12.59} & \multirow{3}{*}{2.19$\times$} & \multirow{3}{*}{1.32$\times$} \\
& & & LN:$24\times\mathbb{R}^{256\times768}$ & & & & & & \\
& & & R:$12\times\mathbb{R}^{256\times3072}$ & & & & & & \\
\midrule
&${\tt SM + ScFuFFN}$ & 3.03 &  SM:$144\times\mathbb{R}^{256\times256}$ & 3.6B & 16.3B & 20.72 & 10.45  & 2.82$\times$  & 1.59$\times$ \\
%&${\tt SM + ScFuFFNi_6}$ & 3.08 & SM:$144\times\mathbb{R}^{256\times256}$ & 1.8B & 16.3B  & 20.59 & 10.32  & 2.84$\times$ & 1.61$\times$ \\ \cline{2-10}
\rowcolor{green!20} &$\text{EReg}({\tt SM(t) + ScFuFFN})$ & 2.92 &  SM:$144\times\mathbb{R}^{256\times256}$ & 3.6B & 16.3B & 20.72 & 10.45    & 2.82$\times$ & 1.59$\times$ \\
%&$\text{EReg}({\tt SM(t) + ScFuFFNi_6})$ & 2.97 & SM:$144\times\mathbb{R}^{256\times256}$ &  1.8B & 16.3B  & 20.59 & 10.32 & 2.84$\times$ & 1.61$\times$ \\
\bottomrule
\end{tabular}}
\end{table}

{\bf GPT-2 Model with 18 Layers}
Table \ref{tab:GPT218LCLen128} provides the latency and communication savings achieved on a 18-layer GPT-2 model, along with a detailed breakdown of the nonlinear operations and FLOPs.

\begin{table}[!h]
\centering
\caption{Results on GPT-2 ($L$=18, $H$=12, $d$=768), trained from scratch on the CodeParrot dataset (2.1B tokens, $T$=128).}
\label{tab:GPT218LCLen128}
\resizebox{0.99\textwidth}{!}{
\begin{tabular}{p{0.2cm}lclcccccc}
\toprule
& \multirow{2}[4]{*}{Network Arch.} & \multirow{2}[4]{*}{PPL} & \multirow{2}[4]{*}{\#Nonlinear Ops} & \multicolumn{2}{c}{\#FLOPs} & \multirow{2}[4]{*}{\shortstack{Comm.\\(GB)}} & \multirow{2}[4]{*}{\shortstack{Lat.\\(min.)}} & \multicolumn{2}{c}{Savings} \\
\cmidrule(lr){5-6} \cmidrule(lr){9-10}
& & & & FFN & Attn. & & & Comm. & Lat. \\
\midrule
\multirow{6}{*}{\rotatebox[origin=c]{90}{Baseline}} & \multirow{3}{*}{${\tt SM + LN + G}$} & \multirow{3}{*}{2.56} & SM:$216\times\mathbb{R}^{128\times128}$ & \multirow{3}{*}{21.7B} & \multirow{3}{*}{11.6B} & \multirow{3}{*}{37.17} & \multirow{3}{*}{10.77} & \multirow{3}{*}{1$\times$} & \multirow{3}{*}{1$\times$} \\
& & & LN:$36\times\mathbb{R}^{128\times768}$ & & & & & & \\
& & & G:$18\times\mathbb{R}^{128\times3072}$ & & & & & & \\ \cline{2-10}
& \multirow{3}{*}{${\tt SM + LN + R}$} & \multirow{3}{*}{2.63} & SM:$216\times\mathbb{R}^{128\times128}$ & \multirow{3}{*}{21.7B} & \multirow{3}{*}{11.6B} & \multirow{3}{*}{13.34} & \multirow{3}{*}{8.04} & \multirow{3}{*}{2.79$\times$} & \multirow{3}{*}{1.34$\times$} \\
& & & LN:$36\times\mathbb{R}^{128\times768}$ & & & & & & \\
& & & R:$18\times\mathbb{R}^{128\times3072}$ & & & & & & \\
\midrule
&${\tt SM + ScFuFFN}$ & 3.24 &  SM:$216\times\mathbb{R}^{128\times128}$ & 2.7B & 11.6B & 8.83 & 6.07  & 4.21$\times$  & 1.77$\times$ \\
\rowcolor{green!20} &$\text{EReg}({\tt SM(t) + ScFuFFN})$ & 3.13 &  SM:$216\times\mathbb{R}^{128\times128}$ & 2.7B & 11.6B & 8.83 & 6.07   & 4.21$\times$ & 1.77$\times$ \\
\bottomrule
\end{tabular}}
\end{table}

\section{Broader Impacts and Potential of Entropy-Guided LLM Solutions}

{\bf Entropy-guided framework for tackling softmax-inherent challenges in attention mechanism}
The softmax function, fundamental to transformer-based attention mechanisms, inherently assigns non-zero probabilities to all tokens due to its normalized exponential structure. This characteristic leads to two primary issues inherent to softmax: disproportionate emphasis on specific tokens (known as attention sink) \cite{xiao2024efficient,cancedda2024spectral,gu2024attention}; and non-zero scores for irrelevant tokens (known as attention noise). These challenges can result in undesirable effects such  as  hallucinations \cite{ye2024differential}, outlier activations \cite{hu2024outlierefficient}, and inefficient use of model capacity, such as rank collapse \cite{bao2024self}. 

While prior research has proposed various strategies to mitigate these issues \cite{yin2024stablemask,yu2024unveiling,bao2024self}, we introduce a principled approach to control attention entropy distribution. By penalizing excessively high entropy values and incorporating learnable threshold parameters, each attention head adaptively determine its optimal degree of focus. This could prevent the over-diffusion of attention scores while {\em preserving} the mathematical properties of softmax. 

{\bf Entropy-guided framework for uncertainty estimation and mathematical reasoning}
Recent progress in entropy-based methodologies, such as the Entropix framework for entropy-guided sampling \cite{entropix2024}, and the discovery of entropy neurons that regulates uncertainty in next-token prediction \cite{gurnee2024universal,stolfo2024confidence}, highlights a major shift in the entropy-drive LLM solutions. These approaches are particularly relevant to improving token-level performance in mathematical reasoning tasks \cite{Langlais2024}. Moreover, the recent FrontierMath benchmarks \cite{glazer2024frontiermath}, which have attracted considerable attention in the research community, further highlight the critical need to improve the reasoning capabilities of LLMs.

A key observation from this research is that the demands of mathematical operations vary in terms of token selection confidence. For instance, deterministic (low-entropy) token selection may be more appropriate for simple arithmetic, while exploratory (high-entropy) token selection may be advantageous in complex problem-solving scenarios.

While our current work focuses on entropy regularization of  attention scores in MHA, this concept can be extended to guide token selection during inference. This is analogous to adaptive temperature strategies, where model creativity is modulated based on logit entropy \cite{velivckovic2024softmax}. Furthermore, controlled entropy pathways tailored to numerical computations, coupled with task-specific entropy thresholds, present a promising direction for future work.

To complement these strategies, our simplified architecture could incorporate reasoning tokens, inspired by pause tokens proposed for reasoning processes \cite{Langlais2024}. By leveraging entropy regularization to influence the model's interaction with such tokens, it is possible to construct more structured and interpretable pathways for mathematical reasoning.

{\bf Parallels between entropy-guided attention and differential attention mechanism}
Our entropy regularization framework exhibits conceptual alignment with the recently introduced Differential Transformer architecture \cite{ye2024differential}, despite notable differences in their methodologies.

Similar to how the differential attention mechanism suppresses attention noise via contrastive learning, entropy-guided attention can achieve comparable outcomes by penalizing excessive dispersal of attention across tokens. Both methods ultimately encourage sparse attention patterns: the Differential Transformer accomplishes this by leveraging differences between attention maps, whereas entropy regularization explicitly penalizes high-entropy attention distributions.

These parallels highlight that selective attention can be fostered through either architectural innovations or targeted regularization strategies. Together, they offer complementary approaches to achieving the shared objective of promoting more focused and efficient attention mechanisms.

\end{document}